%% file: main.tex
\documentclass[10pt,a4paper]{dukenlp}
\usepackage[numbers]{natbib}

\usepackage{helvet,xurl}

\usepackage{amsmath,amsthm}
\usepackage{graphicx,booktabs,tabularx,placeins,flafter,capt-of,wrapfig,colortbl}
\usepackage{xcolor,tikz}
\usetikzlibrary{arrows.meta,positioning,calc}
\usepackage{hyperref}
\definecolor{inkblue}{HTML}{294D70}
\definecolor{contentblue}{HTML}{377EB8}
\definecolor{roleorange}{HTML}{CB7838}

\hypersetup{pdfauthor={Yufa Zhou},pdftitle={The Geometry of Contextual Relations: Language Models Address Facts by Order of Mention}}

\usepackage[capitalize,noabbrev]{cleveref}

\newcommand{\teaserwidth}{.80\linewidth}

\theoremstyle{definition}
\newtheorem{definition}{Definition}[section]
\newtheorem{hypothesis}[definition]{Hypothesis}
\theoremstyle{plain}
\newtheorem{lemma}[definition]{Lemma}
\newtheorem{proposition}[definition]{Proposition}
\newtheorem{corollary}[definition]{Corollary}
\theoremstyle{remark}
\newtheorem{remark}[definition]{Remark}
\crefname{hypothesis}{Hypothesis}{Hypotheses}
\newcommand{\R}{\mathbb{R}}

\title{The Geometry of Contextual Relations: Language Models Address Facts by Order of Mention}

\author{
    \textbf{Yufa Zhou}
    \\
    Duke University\\
    \texttt{yufa.zhou@duke.edu}
}

\date{}
\begin{document}
\addtocontents{toc}{\protect\setcounter{tocdepth}{-1}}
\maketitle
\begin{abstract}
\input{0_abstract}
\end{abstract}
\input{figures/teaser-block}
\input{1_intro}
\input{2_related}

\input{3_geometric_hypothesis}
\input{4_order_of_mention}
\input{5_content_selection}
\input{6_address_geometry}
\input{6_transfer_depth}
\input{7_discussion}

\input{7_conclusion}

\bibliographystyle{plainnat}

\bibliography{ref,notes/perspective-references}
\clearpage
\appendix
\addtocontents{toc}{\protect\setcounter{tocdepth}{2}}
\input{appendix/contents}
\section{Setup}
\label{app:setup}
\input{appendix/selection_protocol}
\section{Invariance and Reversal in Two-Fact Contexts}
\label{app:natural_binding}
\input{appendix/natural_binding}
\section{Questions Are Organized by Order of Mention}
\label{app:order_of_mention}
\Cref{sec:interventions} claims that the query-dependent state is set by the queried fact's order of mention, not by the named person, also when one fact has two subjects and when facts have two arguments; this section gives the query-state geometry and the tests behind that claim.
\input{appendix/query_geometry}
\input{appendix/fact_vs_mention}
\input{appendix/position_controls}
\section{The Fact Address Transfers across Contexts}
\label{app:transfer_controls}
\input{appendix/transfer_controls}
\section{Fact Addresses Are Low-Rank}
\label{app:low_rank}
\Cref{sec:address_geometry} claims that fact addresses span a shared low-rank subspace, crowd together in longer lists, are fixed once their fact is read, and predict how person and relation changes combine; this section gives the analyses behind each claim.
\input{appendix/address_subspace}
\input{appendix/composition}
\section{Where and When the Fact Address Is Shared}
\label{app:where_when}
\Cref{sec:transfer_depth} claims that the fact address is shared as a difference only within a depth window, and that it forms during pretraining; this section gives the every-block design and the checkpoint tests.
\input{appendix/depth}
\input{appendix/training}
\section{Generality: Model Scale, Base Weights, and Prose}
\label{app:generality}
\input{appendix/generality}

\end{document}

%% file: 0_abstract.tex
Human reasoning depends on how objects are related within propositions. \textit{How do relations organize the language representations of contextual contents?} We give an LLM a list of facts in its context (e.g., \emph{Alice eats an apple. Bob eats a pear.}) and measure how its hidden state changes when the question switches from what Alice eats to what Bob eats. Averaged over many lists, this change is a steering vector, which we call the \emph{ordinal vector}. It points to a fact by its \emph{order of mention}, the order in which the facts were stated in the context. We find that LLMs represent the fact a question asks about by its order of mention, not by the name the question contains. We state this as the \textit{ordinal addressing hypothesis}: each order of mention has a \emph{fact address} in the model's state, shared by all contexts, and a question moves the state to the fact address of the fact it asks about, while the context supplies what that fact says. Across Qwen, Gemma, and Llama, fact addresses are (1) \emph{ordered by mention}: query states are organized by the order of facts, not of names, even when one fact has multiple subjects; (2) \emph{steerable}: added to a question about the first fact of a new list, the ordinal vector makes the model answer with the second fact of that list; (3) \emph{low-rank}: they span a low-rank subspace in which the first-mentioned fact is the easiest to reach, surprisingly similar to human recall; and (4) \emph{emergent}: they are shared in late-middle layers, hold from 1.5B to 32B parameters, and form early in pretraining. Language models reach a stated fact by where it was mentioned, deepening our understanding of LLM reasoning.
Code is available at \url{https://github.com/MasterZhou1/order-of-mention}.

%% file: figures/teaser-block.tex
\begin{center}
\includegraphics[width=\teaserwidth]{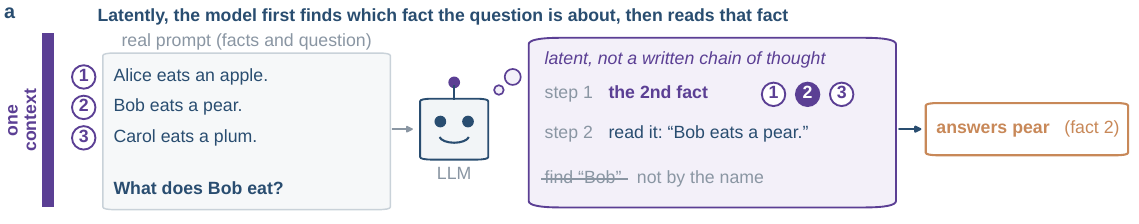}\\[1pt]
\includegraphics[width=\teaserwidth]{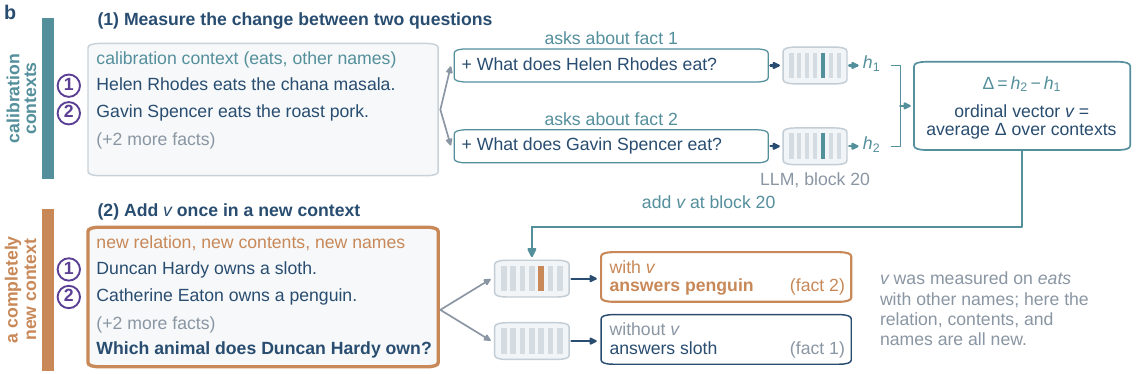}
\captionof{figure}{\textbf{Language models address stated facts by order of mention.} (a) Asked what Bob eats, the model's state after the question points to the second fact, not to the name, and the answer is then read from that fact. (b) A real Qwen2.5-7B run: (1)~the block-20 query change $\Delta$ between questions about fact~1 and fact~2, averaged over calibration contexts, is the ordinal vector $v$; (2)~added once in a new context, $v$ moves the answer to fact~2's animal.}
\label{fig:operations}
\end{center}

%% file: 1_intro.tex
\section{Introduction}
\label{sec:intro}

Human reasoning draws conclusions from relations among objects and events \citep{Halford_2010}. The same inference works for any names, yet its conclusion depends on who plays which role: if Alice gives Bob a book, Bob has it; if Bob gives Alice a book, Alice has it. Combining facts therefore requires preserving each element's semantic role, who does what to whom, even as the contents change \citep{smolensky1990binding}.

Large language models (LLMs) face this problem when reasoning from contextual facts \citep{feng2024bind}.
In \emph{Alice loves Bob. Carol loves David.}, the relation \emph{loves} has two \emph{semantic roles}, the person who loves and the person loved \citep{gildea-jurafsky-2002-automatic}, and a \emph{binding} specifies who fills each role: Bob and David share a role, but whom Alice loves depends on the binding. Semantic structure shapes language-model representations: word-vector analogies and concept geometry reveal directions shared across related meanings \citep{mikolov2013linguistic,park2024linear,park2025categorical}, and contexts trace recurring trajectories as logically structured information accumulates \citep{zhou2025geometry}.
Relations add a particular challenge: the same relational form recurs across examples, while particular facts depend on which objects fill each role. This leads to our central question:
\begin{center}
\emph{How do relations organize the language representations of contextual contents?}
\end{center}

We study it where a question meets the facts. Given the facts \emph{Alice eats the apple} and \emph{Bob eats the pear}, a listener asked what Bob eats can find the fact by its description, the one about Bob, or by its order of mention, the second one, as \emph{the former} and \emph{the latter} do. People track the order in which facts are mentioned, and the first-mentioned participant is the easiest to retrieve \citep{gernsbacher1988accessing,henson1998startend}. Which does a language model use? We change the question from what Alice eats to what Bob eats and measure how the hidden state changes before the model answers. Averaged over contexts, this change is a steering vector, which we call the \emph{ordinal vector} (\cref{fig:operations}).
We find that language models address stated facts by order of mention. We state this as a hypothesis (\cref{sec:geometric_hypothesis}): each order of mention has a \emph{fact address}, shared by all contexts; a question moves the state to the address of the fact it asks about, and the context supplies what is stated there, the fact's \emph{binding identity}. Its consequences hold in Qwen, Gemma, and Llama:

\textbf{(1) Questions are organized by order of mention.} Query changes keep their direction on new names and reverse when the facts are reversed, in fourteen LLMs with base as well as instruction-tuned weights. Query states group by the queried fact's order of mention, not by the named person, and follow the order of facts, not of names, when one fact has two subjects (\cref{sec:geometric_hypothesis,sec:interventions}).

\textbf{(2) The fact address transfers across contexts.} Added once to a context with new names, contents, and relation, the ordinal vector selects the designated fact among four, and that context supplies the content, also in short paragraphs and from 1.5B to 32B parameters (\cref{sec:selection}; \cref{tab:scale}). The vector that transfers follows order of mention, not names, relation, or token distance (\cref{sec:selection}).

\textbf{(3) Fact addresses are low-rank.} They span a low-rank subspace where ordinal vectors compose; later addresses crowd together and the first is easiest to reach (\cref{sec:address_geometry}).

\textbf{(4) The fact address is shared in a depth window and learned early in pretraining.} In late-middle blocks, ordinal vectors and states copied from another context select the receiving context's facts; later states carry the answer's content. In OLMo-2-7B checkpoints, questions are first organized as if by the named person and switch to order of mention by 42B tokens, before the model answers reliably, and the full depth profile forms late (\cref{sec:transfer_depth}).

A question thus says where to look, and the context says what is there, a division learned from next-token prediction alone.

%% file: 2_related.tex
\section{Related Work}
\label{sec:related}

\paragraph{Mechanistic interpretability.}
Circuits, features, and causal interventions explain behavior through internal components and the variables they carry \citep{olah2020zoom,elhage2021mathematical,wang2023interpretability,bricken2023monosemanticity,meng2022locating,geiger2021causal}, and activation addition and function vectors show that an activation estimated across examples can alter behavior on new inputs \citep{rimsky2024steering,todd2024function}. Closest to our question, studies of contextual binding find binding identifiers, propositional probes, and positional and lexical contributions to bound-entity retrieval \citep{feng2024bind,feng2025propositional,prakash2026lookbacks,gurarieh2026mixing}, and editing low-rank subspaces of bound-token states changes bindings \citep{dai-etal-2024-representational,dai2026cell}. \citet{prakash2026lookbacks} find ordering IDs that a lookback mechanism uses to retrieve bound information, and \citet{gurarieh2026mixing} find that this positional mechanism grows noisy in the middle of long lists, where lexical and reflexive mechanisms take over. These works locate order in the states of the bound tokens within one context; we study the question's state across contexts, find that the address it carries counts facts rather than mentioned entities, state its geometry as a hypothesis, and show that this address transfers to new contexts and crowds in the middle of a list, a geometric account of that noise.

\paragraph{Representation analysis and geometry.}
Linear and categorical representation theories tie semantic contrasts and concept hierarchies to directions \citep{mikolov2013linguistic,arora2018linear,park2024linear,park2025categorical}, language models transform concepts with shared geometry \citep{hu2026shared}, and statistical accounts derive such geometry from co-occurrence and its symmetries \citep{jiang2024origins,karkada2026symmetry,modell2025origins,huh2024platonic}. Geometry formed while reading structured input has been studied for in-context graphs, sentence trajectories, and logical reasoning \citep{park2025incontext,hosseini2023straighten,zhou2025geometry}, and steering can depend on how a context is presented \citep{opielka2026invariance}; the geometry we study is set by the order of the stated facts rather than by word meaning.

%% file: 3_geometric_hypothesis.tex
\section{Order of Mention and Fact Addresses}
\label{sec:geometric_hypothesis}

\begin{figure}[!t]
\centering
\includegraphics[width=\linewidth]{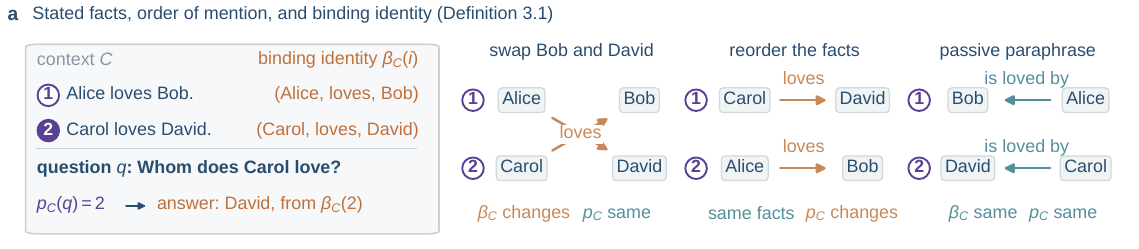}\\[2pt]
\includegraphics[width=\linewidth]{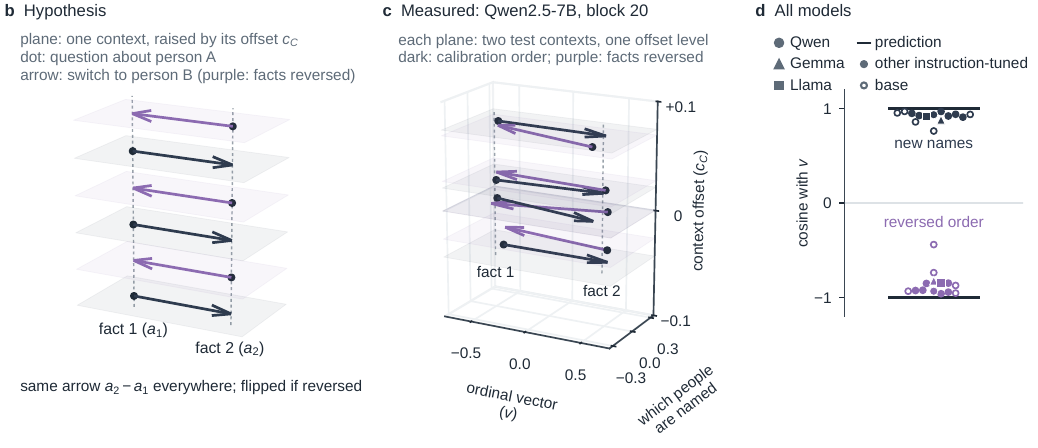}
\caption{\textbf{The problem, the hypothesis, and its first test.} (a) \Cref{def:order_of_mention} on the introduction's example: each stated fact has an order of mention and a binding identity, and a question is answered from the binding identity at its order of mention. Swapping Bob and David changes the binding identities; reordering keeps the facts and changes $p_C$; the passive paraphrase changes neither. (b) \Cref{hyp:conditional_reuse} as a displacement field: every context shares the fact addresses $a_1,a_2$, so the question moves each context by the same $a_2-a_1$, or by $-(a_2-a_1)$ when the two facts are reversed. (c) The same form measured in Qwen for eight test contexts, two arrangements at each of four offset levels (percentiles of the context-offset coordinate), on three axes fit on calibration word sets: along $v$, which people are named, and the context offset; units of $\|v\|$, with the offset axis stretched for display (\cref{app:natural_binding}). (d) Cosines with $v$ for new names (dark; mean of 64 name pairs) and between the ordinal vectors of reversed arrangements (purple), one dot per model, against the predictions of \cref{prop:invariance}; small dots: eleven further models (\cref{app:generality}; open: base weights).
}
\label{fig:relational_hypothesis}
\end{figure}
Given the facts \emph{Alice eats the apple} and \emph{Bob eats the pear}, the question \emph{What does Bob eat?} is answered in two steps: find the fact the question is about, then read it. We formalize the first step.

\begin{definition}[Stated facts, order of mention, and binding identity]
\label{def:order_of_mention}\label{def:binding_identity}
A context $C$ states facts $f_1,\dots,f_n$ in this order, where $f_i$ binds one or more entities $e_i$ and a content $x_i$ under a relation $r_i$. The index $i$ is the fact's \emph{order of mention}, and the \emph{binding identity} at order of mention $i$ is $\beta_C(i)=(e_i,r_i,x_i)$, what $C$ states there. A question $q$ names part of one fact and asks for another part; $p_C(q)$ is that fact's order of mention, and the answer is read from $\beta_C(p_C(q))$. For a fixed relation, $q_C(p)$ is the question that names the entity of fact $f_p$ and asks for its content.
\end{definition}

A question names an entity, but the model could reach the fact by its order of mention instead, as ordinal reference and human serial memory do \citep{henson1998startend}.

\paragraph{Contexts and states.}
A \emph{word set} is one set of names and contents; a context states one fact per name in fixed sentence frames, and its \emph{arrangement} orders the facts while its \emph{content assignment} pairs names with contents. For a context $C$ and question $q$, the \emph{query state} $h_\ell(C,q)\in\R^{d}$ is the residual stream at the last prompt token after Transformer block $\ell$.

\begin{definition}[Query change and ordinal vector]
\label{def:query_response}
For orders of mention $p\neq p'$ and calibration contexts $\mathcal C_{\rm cal}$ from other word sets, define
\begin{equation}
\Delta_\ell(C;q\!\to\!q')=h_\ell(C,q')-h_\ell(C,q),\qquad
v_\ell^{\,p\to p'}=\mathbb E_{C\in\mathcal C_{\rm cal}}\,\Delta_\ell\big(C;q_C(p)\!\to\!q_C(p')\big).
\label{eq:query_mean}
\end{equation}
The mean $v$ is a steering vector, the \emph{ordinal vector}. In a \emph{receiver} $C_*$, intervene once:
\begin{equation}
h_\ell(C_*,q_{C_*}(p))\leftarrow h_\ell(C_*,q_{C_*}(p))+v_\ell^{\,p\to p'},
\label{eq:selection_edit}
\end{equation}
keep the question, and ask whether the complete answer is the \emph{designated answer}, the content of the \emph{designated fact} at $p'$ in $C_*$. Replacing $v$ by the receiver's own $\Delta_\ell(C_*;q_{C_*}(p)\to q_{C_*}(p'))$ gives the \emph{ceiling}.
\end{definition}

\begin{hypothesis}[Ordinal addressing]
\label{hyp:conditional_reuse}\label{def:position_model}
At an intermediate block $\ell$ there are context offsets $c_C$ and \emph{fact addresses} $a_1,a_2,\dots$, one per order of mention and shared by all contexts, such that
\begin{equation}
h_\ell(C,q)=c_C+a_{p_C(q)}+\varepsilon_\ell(C,q)
\label{eq:position_model}
\end{equation}
with small $\varepsilon_\ell$ (measured by held-out $R^2$), and, beyond block $\ell$ (the \emph{readout clause}), the answer is the content of $\beta_C(p)$ when the last-token state is $c_C+a_p$: a question moves the state to the address of the fact it asks about, and the context supplies the binding identity found there.
\end{hypothesis}

Let $y(C,q)=h_\ell(C,q)-\frac1n\sum_{p}h_\ell(C,q_C(p))$ be the \emph{query-dependent state}: the query state centered over the context's $n$ questions, one per fact. With $\varepsilon=0$, the hypothesis has four consequences (\cref{prop:invariance,prop:organization,prop:transfer}, \cref{cor:linear}), each tested below; \cref{prop:margin} relaxes $\varepsilon=0$.

\begin{proposition}[Invariance and reversal]
\label{prop:position_predictions}\label{prop:invariance}
For any contexts $C,C_*$ and orders of mention $p\neq p'$, $\Delta_\ell(C;q_C(p)\to q_C(p'))=\Delta_\ell(C_*;q_{C_*}(p)\to q_{C_*}(p'))=a_{p'}-a_p$, for any names and contents they state. If $C''$ exchanges the facts at $p$ and $p'$ of $C$, the same pair of questions gives $-(a_{p'}-a_p)$.
\end{proposition}

\begin{proposition}[Organization by order of mention]
\label{prop:organization}
$y(C,q)=a_{p_C(q)}-\bar a$ with $\bar a=\frac1n\sum_p a_p$: the query-dependent state depends on $q$ only through $p_C(q)$, not on the entities or relation it names or on token distances. If $C'$ moves the fact at $p$ in $C$ to $\pi(p)$, then $y(C',q)=a_{\pi(p_C(q))}-\bar a$.
\end{proposition}

\begin{proposition}[Transfer]
\label{prop:transfer}
For any receiver $C_*$, \cref{eq:selection_edit} gives $h_\ell(C_*,q_{C_*}(p))+v_\ell^{\,p\to p'}=c_{C_*}+a_{p'}$, so by the readout clause the answer is the content of $\beta_{C_*}(p')$.
\end{proposition}

\begin{corollary}[Linear structure]
\label{cor:linear}
Every change $\Delta_\ell(C;q\to q')$ lies in the address subspace $A_n=\mathrm{span}\{a_p-a_1: p\le n\}$, of dimension at most $n-1$, shared by all contexts of $n$ facts, and ordinal vectors compose: $v^{p\to p''}=v^{p\to p'}+v^{p'\to p''}$.
\end{corollary}
\begin{proof}[Proof of \cref{prop:invariance,prop:organization,prop:transfer} and \cref{cor:linear}]
Substitute \cref{eq:position_model}: $c_C$ cancels in every within-context difference, so every $\Delta_\ell(C;q_C(p)\to q_C(p'))$ and every mean $v^{p\to p'}$ equals $a_{p'}-a_p$; exchanging or moving facts changes only which address each question reaches; the readout clause gives the answer; and differences of addresses lie in $A_n$ and telescope.
\end{proof}

\begin{remark}
A relation component, shared by a context's questions, cancels in $\Delta_\ell$; coding the question by the named entity or the answer's content would contradict \cref{prop:invariance,prop:organization,prop:transfer}.
\end{remark}

\begin{proposition}[Spacing bounds transfer]
\label{prop:margin}
Let instead $\varepsilon_\ell\sim\mathcal N(0,\sigma^2 I)$, let $v$ be exact, $v=a_{p'}-a_p$, and let the readout return the fact whose address is nearest to $h_\ell-c_{C_*}$. The edit of \cref{prop:transfer} then selects $\beta_{C_*}(p')$ with probability between $1-(n-1)\,\bar\Phi(d_{p'}/2\sigma)$ and $1-\bar\Phi(d_{p'}/2\sigma)$, both rising with $d_{p'}=\min_{p''\neq p'}\|a_{p''}-a_{p'}\|$, where $\bar\Phi$ is the standard normal tail: isolated addresses are reached more reliably. This extends the readout clause, to which it reduces at $\varepsilon=0$.
\end{proposition}
\begin{proof}
The edited state is $c_{C_*}+a_{p'}+\varepsilon_\ell$. It is read as fact $p''$ only if the component of $\varepsilon_\ell$ toward $a_{p''}$ exceeds $\|a_{p''}-a_{p'}\|/2$, with probability $\bar\Phi(\|a_{p''}-a_{p'}\|/2\sigma)\le\bar\Phi(d_{p'}/2\sigma)$; a union bound over the $n-1$ other addresses gives the lower bound. A component toward the nearest address beyond $d_{p'}/2$ alone causes an error, which gives the upper bound.
\end{proof}

\paragraph{Models and measurement.}
We use Qwen2.5-7B-Instruct, Gemma-3-4B-IT, and Llama-3.1-8B-Instruct. The \emph{main blocks} are 20, 22, and 15, fixed on separate word sets before testing. Each edit occurs once, before the first answer token, in native chat templates (\cref{app:selection_protocol}). Tests use 64 new word sets per model, and calibration uses others. We score complete greedy answers, counting invalid and truncated outputs as failures.

\paragraph{Invariance and reversal hold across models.}
In two-fact contexts (\cref{app:natural_binding}), the query change behaves as \cref{prop:invariance} predicts (\cref{fig:relational_hypothesis}c,d). Over 64 new name pairs, its mean cosine with the ordinal vector is $0.93$, $0.88$, and $0.92$ in Qwen, Gemma, and Llama, and the ordinal vectors of the two arrangements have cosines of $-0.93$, $-0.82$, and $-0.85$. Eleven further models show the same signs; the weakest reversal cosine is $-0.44$, in base Gemma-3-4B (\cref{fig:relational_hypothesis}d).

%% file: 4_order_of_mention.tex
\section{Questions Are Organized by Order of Mention}
\label{sec:interventions}\label{sec:transfer_conditions}

We next test \cref{prop:organization}: the query-dependent state is set by the queried fact's order of mention.

\begin{figure}[!ht]
\centering
\includegraphics[width=\linewidth]{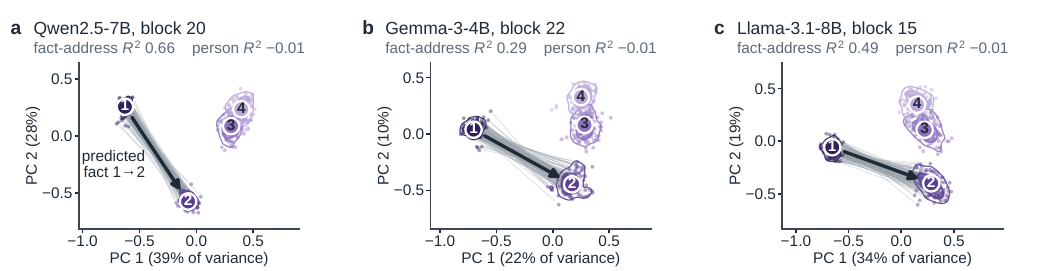}
\caption{\textbf{Query states are organized by order of mention, not by the queried person.} (a--c) Query-dependent states of held-out word sets at each model's main block, projected onto the top two principal directions of the four fact-address templates fit on eight other word sets; axis labels give the share of held-out query-dependent variance on each direction, and the unit is the fact-1 to fact-2 template distance. Each dot is one context under one question, colored by the queried fact's order of mention; gray lines join a context's fact-1 and fact-2 questions, the dark arrow is the predicted change $a_2-a_1$, and contours enclose 50\% and 85\% of each order's density. Panel headers: held-out $R^2$ of fact-address and person templates.
}
\label{fig:query_organization}\label{fig:selection_organization}
\end{figure}

\paragraph{Query states group by the queried fact.}
A \emph{fact-address template}, the mean query-dependent state over calibration contexts whose queried fact has a given order of mention, explains $66\%$, $29\%$, and $49\%$ of the query-dependent variance of held-out word sets in Qwen, Gemma, and Llama, and a \emph{person template}, by the queried person (the $k$-th name of each word set), explains none (\cref{fig:query_organization}). Inside one word set, where the names are fixed, the named person sets the state in early blocks but adds at most $0.10$ to the fact-address template's $R^2$ at the main blocks (\cref{app:within_world}, \cref{fig:within_world}).

\paragraph{The address belongs to the fact, not to the name.}
When one fact has two subjects, the order of facts and the order of names come apart: Carol is the third name but the subject of the second fact in \emph{Alice and Bob eat the apple. Carol eats the pear.} Such query states lie closer to the fact-address template of their fact than to the template at their name's order in $99.2\%$, $96.2\%$, and $96.2\%$ of cases in Qwen, Gemma, and Llama (\cref{app:fact_vs_mention}). In two-place facts such as \emph{Alice loves Bob}, query states are organized by fact and \emph{argument slot}, which argument is asked for, not by person: a fact-by-slot template has held-out $R^2$ of $0.82$, $0.65$, and $0.70$ (\cref{app:role_address}, \cref{fig:role_address}).

%% file: 5_content_selection.tex
\section{The Fact Address Transfers across Contexts}
\label{sec:selection}\label{sec:mean_transfer}
\input{figures/selection-evidence-table}

If the address is shared, it should also move between contexts: by \cref{prop:transfer}, an ordinal vector calibrated on other contexts should select the designated fact in a new context, which supplies its content. With two facts any change reaches the only other content, so we use four and the twelve directed changes between their orders of mention, each averaged over calibration contexts (\cref{app:selection_protocol}). In a list that begins \emph{Petra eats the buckwheat} and \emph{Freya eats the amaranth}, $v^{1\to2}$ added to the question about Petra should give \emph{amaranth}.

\paragraph{The answer follows the receiver's facts.}
Across 64 new word sets, the ordinal vector selects the designated fact in $84.1\%$ of cases in Qwen, $53.9\%$ in Gemma, and $36.2\%$ in Llama (\cref{tab:contextual_selection}A), and equally long random directions in at most $0.01\%$ (\cref{app:transfer_controls}). It still works in short paragraphs that state the facts with varied verbs among unrelated sentences ($44.5\%$, $42.4\%$, and $14.0\%$; \cref{app:prose}).

\paragraph{Facts define the vector.}
With every direction scaled to one length, the ordinal vector calibrated with the facts selects the designated answer in $71.3\%$, $41.7\%$, and $10.3\%$ of cases. Directions computed from inputs without the facts select none from the question alone and at most $3.9\%$ from a list of the names (\cref{tab:wording_controls}B): the vector arises from reading the question against the facts.

\paragraph{A whole state from another context.}
A stronger test replaces the receiver's query state for question $s$ with a \emph{donor} context's state. If the donor asked question $t$ about another fact, receivers answer with their own fact at the order of mention of question $t$ in $80.9\%$, $72.0\%$, and $51.3\%$ of cases, against $1.7\%$, $13.9\%$, and $6.1\%$ when the donor asked the receiver's own question (\cref{tab:contextual_selection}C). It rarely yields the donor's complete answer (\cref{app:all_block_donors}): the state carries which fact to read, not what the donor read.

\begin{figure}[!t]
\centering
\includegraphics[width=\linewidth]{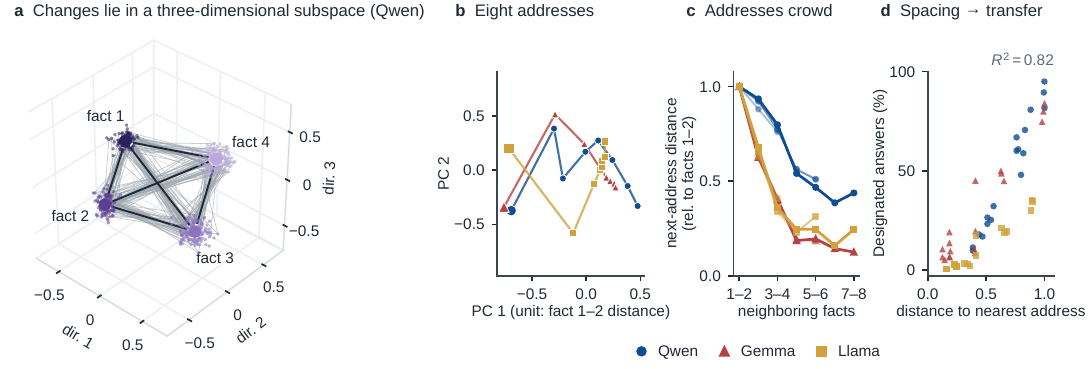}
\caption{\textbf{Fact addresses are low-rank and ordered by mention.} (a) Qwen's held-out contexts in the address subspace fit on eight other word sets: query states (small dots, colored by the queried fact) and changes (gray); large dots: the four fact addresses, joined as a tetrahedron. (b) The eight fact addresses of eight-fact lists in their first two principal directions (unit: fact-1 to fact-2 distance); large marker: fact~1. (c) Distance between neighboring fact addresses, relative to facts 1--2, for four, six, and eight facts (light to dark). (d) Each address's distance to its nearest neighbor (unit: fact-1 to fact-2 distance) against transfer to that fact (\cref{prop:margin}), for 54 addresses; $R^2$ after removing each model's mean.}
\label{fig:address_linear}
\end{figure}
\paragraph{Which ordinal vector transfers.}
It is the one chosen by the receiver's orders of mention, not by its names, relation, or token distances. When one fact has two subjects, a vector chosen by the order of facts selects the designated answer in $65.3\%$ of Qwen cases, and one chosen by the order of names in none (\cref{app:fact_vs_mention}). When the facts are rearranged, the vector of the original question pair mostly fails, and the one chosen by the new orders of mention raises Qwen's transfer for person changes from $9.0\%$ to $69.7\%$ (\cref{app:rearranged}). Across six relations, a vector calibrated only on \emph{eats} and rescaled to the receiving relation's change length comes within five percentage points of each relation's own (\cref{app:relation_generality}). When filler separates order from token distance, the vector selects the fact at the calibrated order over the one at the calibrated absolute distance by $72$ percentage points in Qwen, and by less in Gemma and Llama (\cref{app:address_controls}).

%% file: figures/selection-evidence-table.tex
\begin{table}[t]

\centering\footnotesize
\definecolor{tabfill}{HTML}{548F9C}
\newcommand{\refrow}[1]{\textcolor{black!55}{#1}}
\setlength{\tabcolsep}{2pt}
\begin{minipage}[c]{0.40\linewidth}
\caption{\textbf{An ordinal vector from other contexts selects the designated fact in a new context.} Designated answers (\%), question unchanged; $\pm$ is half the 95\% word-set bootstrap interval; shading is proportional to the rate. (A) 64 new word sets, 12,288 cases per model; Random: at most $0.01\%$. (B) 64 further word sets, 1,024 cases per row, all directions scaled to one length; the question and name rows are computed without the facts (\cref{app:transfer_controls}). Ceiling: the receiver's own state for the designated question; it differs between panels because the word sets differ. (C) The receiver's state replaced by a donor's; 16 donors $\times$ 64 receivers; designated: the receiver's fact at the order of mention of the other question.}
\label{tab:contextual_selection}\label{tab:wording_controls}\end{minipage}\hfill\begin{minipage}[c]{0.56\linewidth}\centering
\begin{tabular}{@{}lrrr}
\toprule
& Qwen & Gemma & Llama\\
\midrule
\multicolumn{4}{@{}l}{\textit{A.\ Transfer to new contexts}}\\
Ordinal vector $v$ & \cellcolor{tabfill!38}\textbf{84.1}{\,\tiny$\pm$2.0} & \cellcolor{tabfill!24}\textbf{53.9}{\,\tiny$\pm$1.6} & \cellcolor{tabfill!16}\textbf{36.2}{\,\tiny$\pm$1.6}\\
\refrow{Random} & \cellcolor{tabfill!0}0.0\phantom{\,\tiny$\pm$0.0} & \cellcolor{tabfill!0}0.0\phantom{\,\tiny$\pm$0.0} & \cellcolor{tabfill!0}0.0\phantom{\,\tiny$\pm$0.0}\\
\refrow{Ceiling} & 96.9{\,\tiny$\pm$1.3} & 96.4{\,\tiny$\pm$1.2} & 91.9{\,\tiny$\pm$1.9}\\
\addlinespace[2pt]
\multicolumn{4}{@{}l}{\textit{B.\ Where the vector comes from}}\\
With facts ($v$) & \cellcolor{tabfill!32}\textbf{71.3}{\,\tiny$\pm$4.0} & \cellcolor{tabfill!19}\textbf{41.7}{\,\tiny$\pm$2.4} & \cellcolor{tabfill!5}\textbf{10.3}{\,\tiny$\pm$2.0}\\
Question alone & \cellcolor{tabfill!0}0.0\phantom{\,\tiny$\pm$0.0} & \cellcolor{tabfill!0}0.0\phantom{\,\tiny$\pm$0.0} & \cellcolor{tabfill!0}0.0\phantom{\,\tiny$\pm$0.0}\\
Names listed, same order & \cellcolor{tabfill!1}2.5{\,\tiny$\pm$1.1} & \cellcolor{tabfill!1}1.3{\,\tiny$\pm$0.8} & \cellcolor{tabfill!2}3.9{\,\tiny$\pm$1.4}\\
Names listed, shuffled & \cellcolor{tabfill!0}0.0\phantom{\,\tiny$\pm$0.0} & \cellcolor{tabfill!0}0.1{\,\tiny$\pm$0.1} & \cellcolor{tabfill!0}0.2{\,\tiny$\pm$0.2}\\
\refrow{Random} & \cellcolor{tabfill!0}0.0\phantom{\,\tiny$\pm$0.0} & \cellcolor{tabfill!0}0.0\phantom{\,\tiny$\pm$0.0} & \cellcolor{tabfill!0}0.0\phantom{\,\tiny$\pm$0.0}\\
\refrow{Ceiling} & 94.4{\,\tiny$\pm$2.4} & 97.9{\,\tiny$\pm$1.1} & 93.2{\,\tiny$\pm$2.4}\\
\addlinespace[2pt]
\multicolumn{4}{@{}l}{\textit{C.\ Whole state from another context}}\\
Donor asked another fact & \cellcolor{tabfill!36}\textbf{80.9}{\,\tiny$\pm$3.1} & \cellcolor{tabfill!32}\textbf{72.0}{\,\tiny$\pm$2.7} & \cellcolor{tabfill!23}\textbf{51.3}{\,\tiny$\pm$3.1}\\
Donor asked the receiver's question & \cellcolor{tabfill!1}1.7{\,\tiny$\pm$0.9} & \cellcolor{tabfill!6}13.9{\,\tiny$\pm$1.6} & \cellcolor{tabfill!3}6.1{\,\tiny$\pm$1.1}\\
\bottomrule
\end{tabular}
\end{minipage}

\end{table}

%% file: 6_address_geometry.tex
\section{Fact Addresses Are Low-Rank}
\label{sec:address_geometry}
\subsection{Fact addresses span a shared low-rank subspace}
If one address set serves every context, the addresses should span a small shared subspace. As \cref{cor:linear} predicts, the three-dimensional address subspace spanned by fact-address templates from other contexts holds $69\%$, $33\%$, and $52\%$ of the squared norm of held-out changes in Qwen, Gemma, and Llama, within 1--8 points of the best three-dimensional subspace, against $0.1\%$ (about $3/d$) for a random one (\cref{fig:address_linear}a; \cref{app:address_subspace}). The subspaces of six relations nearly coincide (mean cosine of the largest principal angle $0.78$--$0.89$), and ordinal vectors estimated on disjoint word sets compose along a path: $v^{p\to p'}+v^{p'\to p''}$ has cosine $0.99$, $0.97$, and $0.98$ with $v^{p\to p''}$. Fact addresses also predict how person and relation changes interact (\cref{app:composition}).

\subsection{Later addresses crowd together, and the first is easiest to reach}
In lists of up to eight facts, the fact addresses form an ordered chain whose later links shorten (\cref{fig:address_linear}b,c). Each address is fixed once its fact is read: the vector from fact~1 to fact~$p$ keeps its direction when later facts are removed (median cosine $0.99$ with the eight-fact vector; \cref{app:prefixes}), so the direction of $a_p-a_1$ does not depend on the list length. Transfer follows this spacing, as \cref{prop:margin} predicts: the farther a fact address lies from its nearest neighbor, the better an ordinal vector reaches it ($R^2=0.82$ within models; \cref{fig:address_linear}d). Accordingly, transfer is highest to the first fact ($82$--$95\%$ in Qwen), falls toward the middle, and in most lists rises slightly at the last: a serial-position curve with strong primacy, as in human recall (\citealp{henson1998startend}; \cref{app:longer_lists}, \cref{fig:longer_lists}).

%% file: 6_transfer_depth.tex
\section{Where and When the Fact Address Is Shared}
\label{sec:transfer_depth}

\begin{figure}[!t]
\centering
\includegraphics[width=\linewidth]{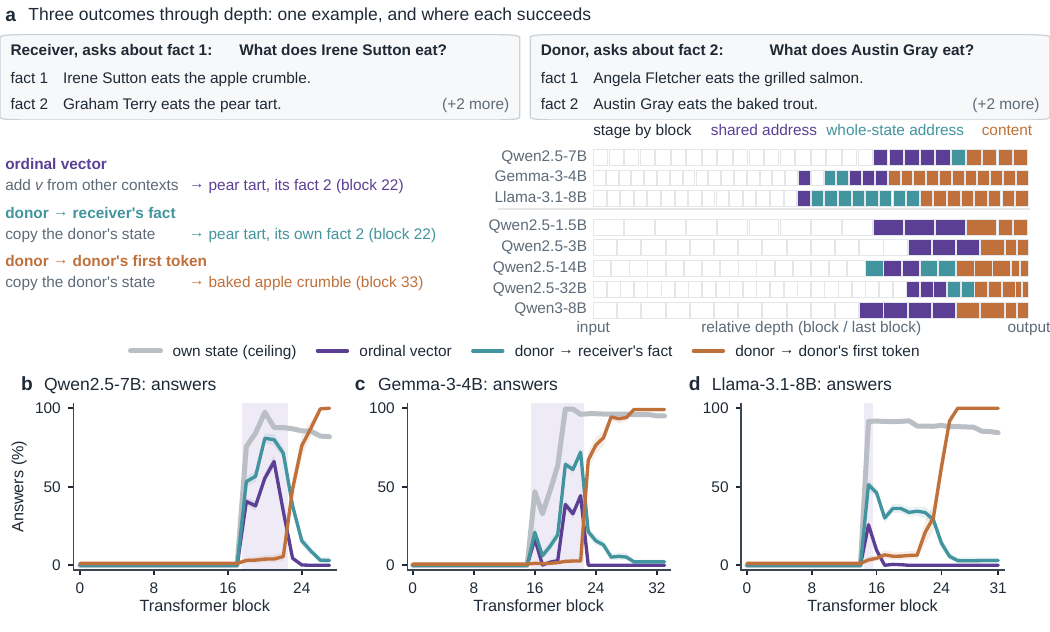}
\caption{\textbf{Through depth, the fact address is first shared across contexts, then held only in whole states, then replaced by the answer's content.} (a) The two interventions and three outcomes on one Gemma example, and each block classified over relative depth: shared address (the ordinal vector selects the receiver's fact in at least $10\%$ of cases), whole-state address (only donor states do, at least $10\%$), and content (the donor's first token reaches $50\%$); blocks are classified one by one, so the first two can interleave; lower rows: other Qwen models (\cref{app:scale}). (b--d) The three outcomes through depth, with the ceiling as reference, on the prompts of \cref{tab:contextual_selection}C (one arrangement, two content assignments, four person changes; 16 donors $\times$ 64 receivers), where the ordinal vector selects $55.5\%$, $44.3\%$, and $25.8\%$ at the main blocks, below \cref{tab:contextual_selection}A over all arrangements; bands: 95\% intervals resampling donor and receiver word sets. Shading: the depth window, blocks where the ordinal vector selects at least $10\%$.}
\label{fig:selection_stages}
\end{figure}

Addressing is an intermediate step: inside one word set, the query state carries first the named person, then the fact address, then the requested content (\cref{app:within_world}). \Cref{hyp:conditional_reuse} should therefore fail near the output, where the state carries the retrieved content $x_p$ of $\beta_C(p)$: what it selects then depends on the context it came from. To see where the address is shared, we apply the two interventions of \cref{sec:selection}, the ordinal vector and the donor state, at every block and record three outcomes: the receiver's fact selected by the ordinal vector, the receiver's fact selected by the donor state, and the donor's own first answer token (\cref{fig:selection_stages}a; \cref{app:all_block_donors}). A state carrying only an address selects the receiver's content; one carrying retrieved content, the donor's.

\begin{figure}[!t]
\centering
\includegraphics[width=\linewidth]{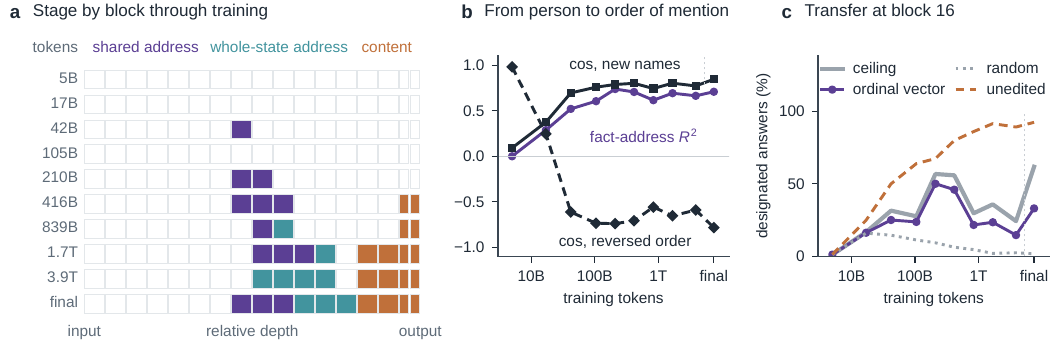}
\caption{\textbf{In pretraining, questions switch from the named person to order of mention before the model answers reliably, and the full depth profile forms late.} Ten checkpoints of OLMo-2-7B, from 5B training tokens to the released model (\emph{final}, after mid-training). (a) Stages as in \cref{fig:selection_stages}a, with each threshold applied to the gain over the unedited answer; all three stages first co-occur at 839B. (b) Fact-address $R^2$ (peak over blocks) and the cosines of \cref{prop:invariance} at block 16. (c) Designated answers at block 16; dashed: unedited accuracy; dotted: random directions.}
\label{fig:development}
\end{figure}
\subsection{The shared address occupies a depth window}
In early blocks even the ceiling has no effect, though order of mention is already readable (\cref{fig:depth_geometry}); in a late-middle \emph{depth window}, ordinal vectors and donor states both select receiver contents (\cref{fig:selection_stages}b--d). The ordinal vector fails first: Gemma's falls from $44.3\%$ at block 22 to zero at block 23 (\cref{fig:selection_stages}c), while the ceiling stays at $96\%$. In Llama, from block 17 to 21, the ordinal vector stays below $1\%$ while donor states select $30$--$36\%$: each whole state holds the address, which is no longer a shared difference. The donor's own first token then takes over, reaching $67.0\%$ at block 23 in Gemma (\emph{baked apple crumble} joins the donor's first word to the receiver's content). The geometry agrees: fact-address $R^2$ collapses where the shared address ends, from $0.29$ at block 22 to $0.05$ at block 23 in Gemma, and each state aligns with its answer's output direction (\cref{fig:depth_geometry}).

\subsection{The fact address forms during pretraining}
\label{sec:development}

We repeat the tests on ten checkpoints of OLMo-2-7B \citep{olmo2024olmo2}, from 5B training tokens to the released model (\cref{fig:development}; \cref{app:development}). Questions switch from the named person to order of mention: at 5B tokens the query change is the same in both arrangements (cosine $+0.98$), as if questions were represented by the named person, and it reverses by 42B ($-0.61$), as \cref{prop:invariance} predicts, while the model answers only half the questions; fact-address $R^2$ rises from $0.00$ to $0.52$ over the same span. The depth profile forms later and settles only at the end of training (\cref{fig:development}a): shared-address blocks first appear at 42B but are absent at 105B and again at 3.9T, content blocks appear from 416B, all three stages first co-occur at 839B, and only the released model shows them as one contiguous profile. Transfer at block 16, the released model's main block, peaks at 210B ($50.1\%$ against a $56.8\%$ ceiling) and then declines as the depth profile moves away from it.

%% file: 7_discussion.tex
\section{Discussion}
\label{sec:discussion}

\paragraph{Order of mention is learned from pretraining.}
Resolving a question by order of mention separates the structure of a context from its contents: one address, which counts facts rather than tokens, serves any names and contents, and the context supplies them. No objective asks for this separation, yet OLMo-2-7B learns it early in pretraining, after first organizing questions as if by the named person. It also shares two signatures of human discourse processing: the first-mentioned fact is the easiest to reach, and later ones crowd together, as in serial recall \citep{gernsbacher1988accessing,henson1998startend}. Unlike linear concept representations \citep{park2024linear}, a fact address is a place in the context, not a meaning.

\paragraph{Fact retrieval as addressing.}
Retrieving a stated fact may then depend on where it stands: crowded addresses in the middle of a list transfer least, which may help explain weaker retrieval in the middle of long contexts \citep{liu2024lost} and reasoning that changes when premises are reordered \citep{chen2024premise}. We study templated lists and short paragraphs and edit one token position at one block. In this setting the address is clearest in Qwen; fact-address templates explain $29\%$ of Gemma's query-dependent variance, and the ordinal vector moves Llama's answers less often in paragraphs and with two-subject facts ($12$--$14\%$). Long lists also draw on lexical and reflexive retrieval \citep{gurarieh2026mixing}. The natural next steps are long documents and multi-step reasoning, where a fact reached by its address must be combined with others.

%% file: 7_conclusion.tex
\section{Conclusion}
\label{sec:conclusion}

We asked how relations organize the language representations of contextual contents, and answered by measuring and editing the state a question leaves before the model answers. Language models address stated facts by order of mention: a question moves the state to a fact address shared across contexts, and the context supplies what is stated there. These addresses steer answers in new contexts, span a low-rank subspace in which later facts crowd together, are shared in a late-middle depth window, and form early in pretraining.

%% file: appendix/contents.tex
\phantomsection
\pdfbookmark[0]{Appendix contents}{appendix.contents}
\section*{Appendix Contents}
The appendix follows the main text. Appendix~\ref{app:setup} gives the shared protocol, and each later part opens with the main-text claim it supports.

\begingroup
\small
\setlength{\parskip}{0pt}
\makeatletter
\@starttoc{toc}
\makeatother
\endgroup
\clearpage

%% file: appendix/selection_protocol.tex
\label{app:selection_protocol}
Every intervention uses one protocol: a change of the query state is calibrated on some word sets, added once to contexts built from new word sets, and judged by the complete answer.

\paragraph{Models and main blocks.}
We use Qwen2.5-7B-Instruct, Gemma-3-4B-IT, and Llama-3.1-8B-Instruct; block indices are zero-based Transformer block outputs. Qwen's main block 20 and Gemma's block 22 were chosen before the experiments reported here, as the blocks where an earlier, related edit of the query state changed the most complete answers on separate word sets, and both were confirmed on the block-selection word sets of this protocol before any test. Llama's block 15, and the main block of every model in \cref{app:generality}, come from the following rule on those block-selection word sets. For each tested block, the rule scores the ordinal vector's gain over the unedited answer, minus the mean gain of random directions, minus any rise in invalid answers; it finds the best block and chooses the earliest block within two points of it. On the blocks profiled for Qwen it also gives block 20; on Gemma's plain-prompt profile it ties blocks 20 and 22 ($51.6$ points each) and would take block 20. The models of \cref{app:generality} test every second block, and where no block reaches two points, the block nearest Qwen2.5-7B's relative depth is used. No main block was chosen on test word sets.

\paragraph{Word sets.}
A \emph{word set} is a set of names and contents: four people and four contents in the four-person experiments. Except where noted, each experiment draws new names and new contents, mostly multiword dishes (\emph{ackee and saltfish}, \emph{pad see ew}). Calibration and test word sets share no complete name or content string, although subwords can recur. Within every word set, the four answers have pairwise distinct first tokens under the model's tokenizer. Every experiment tests on 64 word sets not used for calibration, except the depth curves of \cref{app:generality} and \cref{app:development}, which use 8 donor and 32 receiver word sets. Unless stated otherwise, calibration uses 16 word sets; \cref{tab:contextual_selection}A--B, the two-fact contexts of \cref{app:natural_binding}, and the rearranged facts of \cref{app:rearranged} use 8.

A context states one fact per person, such as \emph{Petra eats the buckwheat}, followed by a question, \emph{What does Petra eat?}, under the instruction \emph{Read the sentences. Answer the question with only the food, in lowercase, with no explanation.} A \emph{content assignment} assigns the four contents to the four people; an \emph{arrangement} orders the four facts. Both rotate cyclically, so every person and every content occupies every order of mention. For \cref{tab:contextual_selection}A, the ordinal vector $v^{p\to p'}$ for a directed pair of orders of mention averages \cref{eq:query_mean} over all calibration contexts, content assignments, and arrangements that realize it. Each test word set crosses four arrangements, four content assignments, and twelve directed changes: 192 cases per word set, 12,288 per model and condition.

\paragraph{Intervention.}
We edit the query state, the state of the last prompt token, at the output of one block, once, while the prompt is processed. All later blocks and positions are recomputed on the receiving context, and generation continues from there. Prompts use each model's native chat template (\cref{app:base_instruct} adds a plain prompt), and computation is in FP32. Random controls are two fixed random directions with the length of the ordinal vector they replace. The \emph{ceiling} replaces only the receiver's last-token state at the edited block with its own state for the designated question $q'$; the question tokens still read $q$, so it measures how far replacing this one state moves the answer.

\paragraph{Decoding and scoring.}
Answers are greedy, with at most 16 new tokens. We score the complete normalized answer string: the \emph{designated} answer is the content that the designated question $q'$ asks for in the receiving context. An answer is \emph{invalid} when its normalized string is not exactly one content of the context. Invalid and truncated answers remain in every denominator, as do cases whose unedited answer is wrong.

\paragraph{Intervals.}
Intervals are 95\% percentile intervals from 5,000 bootstrap draws over test word sets, keeping all content assignments, arrangements, questions, and paired conditions of a word set together. For donor states, donor and receiver word sets are resampled independently. All intervals are pointwise, one per reported rate. In the text, intervals are given in brackets; in tables, $\pm$ gives half the interval. Differences between rates are in percentage points (``points'').

%% file: appendix/natural_binding.tex
\Cref{sec:geometric_hypothesis} claims that in two-fact contexts the query change between the two people's questions keeps its direction under new names and reverses with the order of mention (\cref{prop:invariance}); this section gives the materials, the coordinates of \cref{fig:relational_hypothesis}c, and the cosines of \cref{fig:relational_hypothesis}d.

\paragraph{Materials.}
Each context states two facts, such as \emph{Ruby eats the tuna} and \emph{Simon eats the guava}, and asks \emph{What does Ruby eat?} There are 72 word sets, each with two names and two pairs of contents, and a context uses one pair: 8 word sets calibrate, and 64 are test word sets. All 144 names are distinct, calibration and test contents disjoint, and a word set's contents start with different letters. Each word set is realized in the active form and in the passive form (\emph{The tuna is eaten by Ruby.}), with both content assignments and both arrangements, and the question follows the facts. Every model answers every calibration question correctly.

\paragraph{Coordinates.}
Label the two people $A$ and $B$ within each word set; the labels stay with the people when their sentences are reversed. For arrangement $o\in\{0,1\}$ ($o=0$: $A$'s fact first), let $v_o$ be the mean of $h(C,q_B)-h(C,q_A)$ over both content assignments and the eight calibration word sets, so the ordinal vector is $v=v_0$. \Cref{fig:relational_hypothesis}c uses three axes fit on calibration word sets only: the arrangement contrast $u_1\propto v_0-v_1$, close to $v_0$ because $v_1\approx-v_0$; the component $u_2$ of $(v_0+v_1)/2$ orthogonal to $u_1$ (``which people are named''); and the top principal direction of the calibration context offsets $\tfrac12(h(C,q_A)+h(C,q_B))$ after projecting out $u_1,u_2$. States are centered by the mean calibration offset and expressed in units of $\|v\|$, the root-mean-square norm of $v_0,v_1$. \Cref{fig:relational_hypothesis}c draws the offset axis stretched for display; its tick labels give measured values. The eight displayed test contexts are the active contexts with the first content assignment at the 10th, 37th, 63rd, and 90th percentiles of the offset coordinate, four in each arrangement.

Along $u_2$, test pairs separate by whether a woman or a man is named first, because every calibration pair names a woman first: $u_2$ carries which two people are named, while $u_1$, the arrangement contrast within the same pairs, carries the order of mention. The plane of $u_1,u_2$ holds $91.7\%$, $88.8\%$, and $94.8\%$ of the squared norm of the test changes in Qwen, Gemma, and Llama.

\paragraph{Cosines.}
\Cref{fig:relational_hypothesis}d uses full-dimensional states. For each of the 64 test word sets, the change of that word set in a given arrangement has mean cosine $0.93$, $0.88$, and $0.92$ with the calibration $v_o$ of the same arrangement (minimum $0.85$, $0.71$, and $0.78$). The changes calibrated in the two arrangements are nearly opposite: $\cos(v_0,v_1)=-0.93$, $-0.82$, and $-0.85$. The query change keeps its direction when the names change and flips when the order of mention flips, as a difference of two fact addresses should.

%% file: appendix/query_geometry.tex
\subsection{Query-State Geometry}
\label{app:query_geometry}
Query states are organized by the order of mention of the queried fact (\cref{fig:query_organization}a--c): fact-address templates predict held-out states, and person templates predict nothing.

\paragraph{Data.}
This geometry and that of \cref{fig:depth_geometry} use last-token states saved at every block for the four-person contexts behind \cref{tab:contextual_selection}C. Calibration contexts are 16 word sets $\times$ 4 content assignments $\times$ 4 arrangements, 256 contexts, each with all four questions; because the arrangement rotates, every person occupies every order of mention. Test contexts are the 64 test word sets $\times$ 2 content assignments in one arrangement, 128 contexts. For every block, the query-dependent state is $y(C,q)=h(C,q)-\frac14\sum_{q'}h(C,q')$ at the last prompt token.

\paragraph{Templates and held-out variance.}
A fact-address template $a_p$ averages $y(C,q)$ over calibration contexts by the order of mention $p$ of the queried fact; a person template $b_e$ averages by the queried person $e$. Held-out variance explained is $R^2=1-\sum\|y-a_{p(q)}\|^2/\sum\|y\|^2$, and likewise for $b_e$; a negative value means the template predicts worse than zero. We report the leave-one-word-set-out estimate on calibration contexts, here and at every block of \cref{fig:depth_geometry}: templates fit on 15 word sets predict the held-out word set, and sums are pooled over the 16 held-out word sets. Only calibration contexts can separate the two templates: test contexts keep one arrangement, in which person and order of mention coincide. At the main blocks the estimate is $0.66$, $0.29$, and $0.49$ for the fact-address template and $-0.01$ for person in all three models.

\paragraph{Projection.}
The plane of \cref{fig:query_organization}a--c is spanned by the top two principal directions of the four fact-address templates fit on eight calibration word sets. The points are the query-dependent states of the other eight calibration word sets, unseen by the fit, in all four arrangements and content assignments. Each model contributes 512 points, and the axis labels give the share of their variance on each displayed direction.

\begin{figure}[!ht]
\centering
\includegraphics[width=\linewidth]{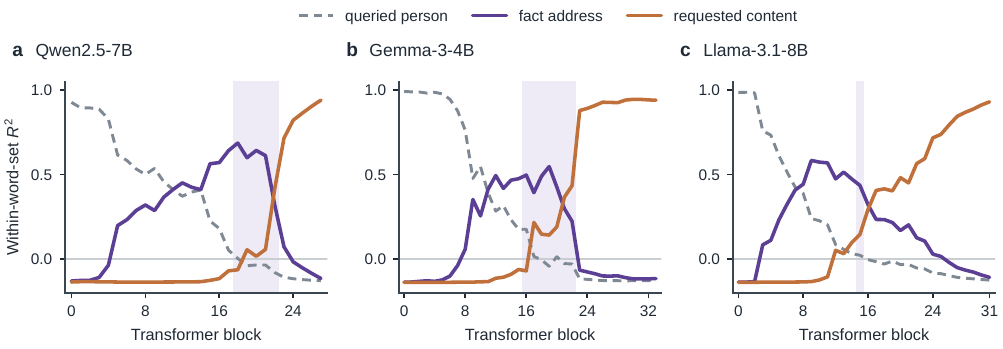}
\caption{\textbf{Inside one word set, the query state passes from the named person to the fact address to the requested content.} Held-out $R^2$ of query-dependent states under templates for the queried person, the queried fact's order of mention, and the requested content, each fit on the other 15 contexts of the same word set and pooled over 16 calibration word sets. Shading: first to last block where the ordinal vector selects at least $10\%$ of designated answers (\cref{fig:selection_stages}).}
\label{fig:within_world}
\end{figure}

\paragraph{Within one word set.}
\label{app:within_world}
Across word sets the names change, so a person template can only carry what the $k$-th names of different word sets have in common. Inside one word set the four names and contents stay fixed while its 16 contexts put every person at every order of mention. For each calibration word set and each held-out context, we fit templates on the other 15 contexts of that word set and score the held-out context's four query-dependent states, pooling over the 16 word sets (\cref{fig:within_world}). At the first block the state is set by the person the question names ($R^2$ $0.93$, $0.99$, and $0.98$ in Qwen, Gemma, and Llama); this share falls to zero as the share of the fact address rises. The fact address peaks inside the window of \cref{fig:selection_stages} in Qwen and Gemma (blocks 18 and 19) and earlier in Llama (block 9), and the requested content, a template that averages by the content the question asks for, takes over in later blocks. At the main blocks the queried person explains $-0.04$, $-0.03$, and $0.02$ of the variance and adds $0.07$, $0.02$, and $0.10$ to a fact-address template, which alone explains $0.64$, $0.22$, and $0.44$. The requested content is already present there in Gemma ($0.43$), whose main block ends its window. At the main blocks the fact address, not the name, sets the state.

%% file: appendix/fact_vs_mention.tex
\subsection{Coordinated Subjects: The Order of Facts, Not of Names}
\label{app:fact_vs_mention}
The address belongs to the fact, not to the name (\cref{sec:interventions,sec:selection}). In every other experiment each fact has one subject, so the $k$-th fact and the $k$-th name coincide. Here one of four facts has two coordinated subjects, \emph{Alice and Bob eat the apple. Carol eats the pear. Dave eats the plum. Emma eats the fig.}, so five names are mentioned in four facts and, after the coordinated fact, a person's name order exceeds the order of their fact by one. Calibration uses ordinary four-fact lists of 16 word sets in four arrangements and four content assignments: fact-address templates $a_p$ of the query-dependent state by the queried fact's order of mention, and the twelve ordinal vectors $v^{p\to p'}$. Test contexts come from 64 new word sets of five people and four multiword contents, with the coordinated fact at each of the four orders of mention and two content assignments, and every single-person question is asked: 2,560 questions, of which 1,280 have a fact order $f$ that differs from their name order $m$.

\paragraph{Geometry.}
For each question with $f\neq m$ and $m\le 4$ (768 questions), the query-dependent state $y$ is compared with the template of its fact order, $a_f$, and with the template of its name order, $a_m$, in the address subspace. It is closer to $a_f$ in $99.2\%$ $[98.6,99.7]$ of questions in Qwen, $96.2\%$ $[94.9,97.5]$ in Gemma, and $96.2\%$ $[94.8,97.5]$ in Llama, and the mean advantage $\cos(y,a_f)-\cos(y,a_m)$ is $0.94$, $0.61$, and $0.68$. The full space gives the same fractions. Held-out calibration questions are assigned to their own template in $100\%$, $100\%$, and $99.4\%$ of cases.

\paragraph{Behavior.}
For receivers whose fact and name orders differ and each target order of mention $t$, an ordinal vector chosen by fact order, $v^{f\to t}$, is compared with one chosen by name order, $v^{m\to t}$. Only targets for which the two orders predict different contents, and for which the name-order prediction is not the original answer, enter the contrast. The fact-order vector selects the content of fact $t$ in $65.3\%$ $[63.0,67.7]$ of cases in Qwen, $34.0\%$ $[32.4,35.6]$ in Gemma, and $12.1\%$ $[10.5,13.7]$ in Llama. The name-order vector selects the content its order predicts in $0.0\%$ of cases in all three models and mostly leaves the original answer. On receivers whose two orders agree, the same vectors select their target in $78.2\%$, $57.7\%$, and $21.9\%$ of cases, and the ceiling in $97.9\%$, $97.1\%$, and $84.7\%$; random directions keep the original answer in at least $99.9\%$ of cases, and unedited answers are correct in at least $99.9\%$. An ordinal vector acts only when it starts at the state's own fact address, so both measures place each question at the address of its fact, not of its name.

%% file: appendix/position_controls.tex
\subsection{Two-Place Relations}
\label{app:role_address}
Queries over two-place facts are organized by fact and argument slot, not by person (\cref{sec:interventions}). Each context states three facts of one two-place relation among six people, such as \emph{Alice loves Bob. Carol loves David. Emma loves Frank.}, with five relations (\emph{loves}, \emph{trusts}, \emph{helps}, \emph{visits}, \emph{pays}). A question asks for one argument given the other, \emph{Whom does Alice love?} or \emph{Who loves Bob?} Every answerable question is a pair of an order of mention and an argument slot, so a query change can move to another fact (same slot), to the other argument of the same fact, or to both. Calibration uses 16 word sets with all arrangements and all content assignments, whose contents are people. The 64 test word sets, with new names whose first tokens are distinct, are edited once at the main block with each calibration mean, the means of the other four relations, two random directions, and the ceiling. Templates are fit on calibration word sets and scored on the test word sets.

\begin{figure}[!ht]
\centering
\includegraphics[width=\linewidth]{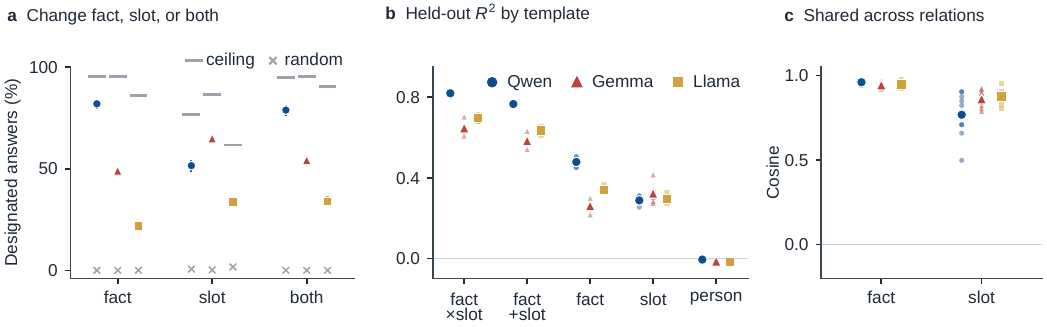}
\caption{\textbf{Queries over two-place facts address a fact and an argument slot.} (a) Designated answers for changes of fact, of argument slot, and of both, against the ceiling (gray ticks) and random directions ($\times$), on 64 new active contexts. (b) Held-out $R^2$ under fact-by-slot, additive (fact$+$slot), fact-only, slot-only, and person templates. (c) Cosines of fact and of slot changes across relations. In (b) and (c), small markers: individual relations; large markers: their mean.}
\label{fig:role_address}
\end{figure}

All three changes transfer to new names: fact, slot, and joint changes select the designated person in $82\%$, $52\%$, and $79\%$ of Qwen cases, $49\%$, $65\%$, and $54\%$ of Gemma cases, and $22\%$, $34\%$, and $34\%$ of Llama cases, while random directions stay at or below $1.7\%$ (\cref{fig:role_address}a). For fact changes, a change calibrated on the other four relations comes within four points of the receiving relation's own. Query states are organized by the two coordinates: a fact-by-slot template explains $0.82$, $0.65$, and $0.70$ of held-out variance, an additive template $0.77$, $0.58$, and $0.63$, and the queried person nothing (\cref{fig:role_address}b). Fact changes estimated on different relations point the same way (mean cosine $0.94$--$0.96$), and so do slot changes ($0.77$--$0.88$). A question about a two-place fact addresses the fact by its order of mention and, within it, the argument slot it asks for.

%% file: appendix/transfer_controls.tex
\Cref{sec:selection} claims that an ordinal vector from other contexts selects the designated fact in a new context, that the facts define this vector, that a whole donor state carries only which fact to read, and that the vector that transfers is chosen by the receiver's orders of mention, not by its names, relation, or token distances. Random directions of each ordinal vector's length select the designated fact in at most $0.01\%$ of the 12,288 cases per model (\cref{tab:contextual_selection}A). We give the controls of \cref{tab:contextual_selection}B--C, then which vector transfers under rearranged facts, six relations, and token distance.

\subsection{What the Facts and Whole Donor States Contribute}
\label{app:transfer_facts_donors}

\paragraph{No-facts directions.}
\Cref{tab:contextual_selection}B compares directions from different inputs at one length. The ordinal vector for a directed pair of orders of mention is the calibration mean of \cref{eq:query_mean} over eight calibration word sets, four arrangements, and four content assignments (128 query changes per pair). Each no-facts direction is the difference between the states for the two questions of the case, computed on an input without the facts: the question alone; the receiving names listed in the order of their facts; or the same names in a shuffled order. Every direction, including the random ones, is scaled to one common length, the root-mean-square length of the twelve factual calibration directions. Each of 64 test word sets contributes two arrangements, two content assignments, and four directed changes, 1,024 cases per row. These word sets and this subset of arrangements and content assignments differ from those of \cref{tab:contextual_selection}A, so the two panels' rates are not directly comparable. The no-facts inputs contain the question words and the names but no facts, and their directions select at most $3.9\%$ of designated answers.

\paragraph{Whole donor states.}
In \cref{tab:contextual_selection}C, the donors are the 16 calibration word sets and the receivers the 64 test word sets, all in one fixed arrangement. For each donor, receiver, content assignment, and directed person change $s\to t$, the receiver asks question $s$, and its state for $s$ is replaced by the donor's state for question $t$ (another fact) or for question $s$ (the receiver's own question); in both rows the designated answer is the receiver's fact at $t$, which gives 8,192 cases per condition. The paired gain of the first replacement over the second is $79.1$ $[75.4,82.6]$, $58.1$ $[54.4,61.9]$, and $45.2$ $[42.0,48.2]$ percentage points, and it is positive for every donor: the smallest per-donor gain is $68.0$, $50.2$, and $38.5$ points. On the same prompts, which are also those of the depth curves of \cref{fig:selection_stages}, the ordinal vector selects $55.5\%$, $44.3\%$, and $25.8\%$, below the whole donor state and below its rates in \cref{tab:contextual_selection}A ($84.1\%$, $53.9\%$, and $36.2\%$), which average over all arrangements. Complete donor answers are rare at the main blocks (\cref{fig:selection_stages}), and at the final block the same replacement gives them in only $2.1\%$, $8.2\%$, and $3.6\%$ of cases, with $91.0\%$, $87.1\%$, and $79.5\%$ invalid (\cref{app:all_block_donors}). A donor state carries which fact to read, not the donor's content.

\subsection{Rearranged Facts}
\label{app:rearranged}
When facts move, the ordinal vector must follow their new orders of mention (\cref{sec:selection}). Each context states four facts about two people and two relations, (\emph{Alice eats the apple. Alice carries the pear. Bob eats the plum. Bob carries the peach.}), and a question asks about one of the four \emph{cells}, a pair of a person and a relation; a person change keeps the relation, and a relation change keeps the person. Calibration uses eight word sets in the person-major arrangement above (grouped by person). The test contexts, on 64 word sets, state the facts in two held-out arrangements, $(1,4,2,3)$ and $(3,2,4,1)$, numbering the facts of the person-major arrangement, with four content assignments and four starting questions: 2,048 cases per condition and change type. Three edits target the same designated answer. The \emph{original-pair} edit keeps the ordinal vector for the same pair of questions, which in the new arrangement connects different orders of mention. The \emph{new-order-of-mention} edit takes the ordinal vector for the pair of orders of mention that the two questions occupy in the new arrangement. The \emph{recalibrated} edit is estimated on calibration word sets in the held-out arrangement.

\begin{table}[!ht]
\centering\small
\caption{\textbf{When facts move, the ordinal vector must follow their orders of mention.} Designated answers (\%) under two held-out arrangements, 64 test word sets and 2,048 cases per entry; $\pm$ is half the 95\% word-set bootstrap interval; bold: the new-order-of-mention edit; ceiling: the receiver's own state for the designated question; no facts: the direction computed from the question alone.}
\label{tab:rearranged}
\vspace{4pt}
\setlength{\tabcolsep}{3pt}
\begin{tabular}{llrrrrrr}
\toprule
Model & Change & Original pair & \shortstack{New order\\of mention} & Recalibrated & Ceiling & Random & No facts\\
\midrule
Qwen & Person & 9.0{\,\tiny$\pm$1.1} & \textbf{69.7}{\,\tiny$\pm$2.3} & 80.9{\,\tiny$\pm$2.5} & 90.2{\,\tiny$\pm$1.7} & 0.0 & 0.0\\
 & Relation & 22.6{\,\tiny$\pm$1.6} & \textbf{61.6}{\,\tiny$\pm$2.1} & 68.5{\,\tiny$\pm$1.9} & 77.3{\,\tiny$\pm$1.9} & 5.7 & 7.0\\
\addlinespace[2pt]
Gemma & Person & 17.6{\,\tiny$\pm$0.9} & \textbf{40.8}{\,\tiny$\pm$1.4} & 61.1{\,\tiny$\pm$2.1} & 79.2{\,\tiny$\pm$1.7} & 0.1 & 0.1\\
 & Relation & 5.5{\,\tiny$\pm$1.3} & \textbf{19.9}{\,\tiny$\pm$2.9} & 41.5{\,\tiny$\pm$2.8} & 73.6{\,\tiny$\pm$1.6} & 0.8 & 0.7\\
\addlinespace[2pt]
Llama & Person & 6.2{\,\tiny$\pm$1.4} & \textbf{15.7}{\,\tiny$\pm$2.8} & 16.4{\,\tiny$\pm$2.8} & 29.8{\,\tiny$\pm$3.7} & 0.0 & 0.0\\
 & Relation & 0.7{\,\tiny$\pm$0.5} & \textbf{0.6}{\,\tiny$\pm$0.3} & 2.8{\,\tiny$\pm$1.1} & 11.8{\,\tiny$\pm$2.1} & 0.3 & 0.3\\
\bottomrule
\end{tabular}
\end{table}

Choosing the ordinal vector by the new orders of mention recovers most of the recalibrated rate in Qwen and nearly all of it for Llama's person changes, and it more than doubles Gemma's rates while leaving a larger gap to recalibration (\cref{tab:rearranged}). Llama's relation changes do not act in these contexts even when recalibrated, and even the ceiling reaches only $11.8\%$. In Qwen, unedited answers to \emph{eats} questions sometimes name the same person's carried item, so $6.4\%$ of relation-change cases already give the designated answer before any edit; random and no-facts directions leave that rate unchanged (gains of $-0.7$ and $+0.5$ points), which is why their relation row is not zero.

\subsection{Six Relations}
\label{app:relation_generality}
The same ordinal vector acts under every relation (\cref{sec:selection}). Each context lists four facts under one of six relations, each with its own contents: \emph{eats} a dish, \emph{owns} an animal, \emph{visited} a city, \emph{works as} a job, \emph{plays} an instrument, or \emph{plays} a sport. The question asks for one person's content under the same relation, and the model answers with only that content. Each relation has 16 calibration and 64 test word sets, with new names and contents in the test word sets. Calibration follows \cref{sec:selection}: for each of the twelve directed changes between orders of mention, the mean query change over a relation's calibration contexts at the main block. Each test receiver is edited once with the ordinal vector of every relation, with the \emph{eats} vector rescaled to the receiving relation's own change length, with two random directions of that length, and with the ceiling. Other cross-relation changes keep the source relation's length, so their gap to the receiving relation's own change mixes direction and length. Geometry uses the query-dependent states at this block, with templates fit on calibration and tested on test word sets.

\begin{figure}[!ht]
\centering
\includegraphics[width=\linewidth]{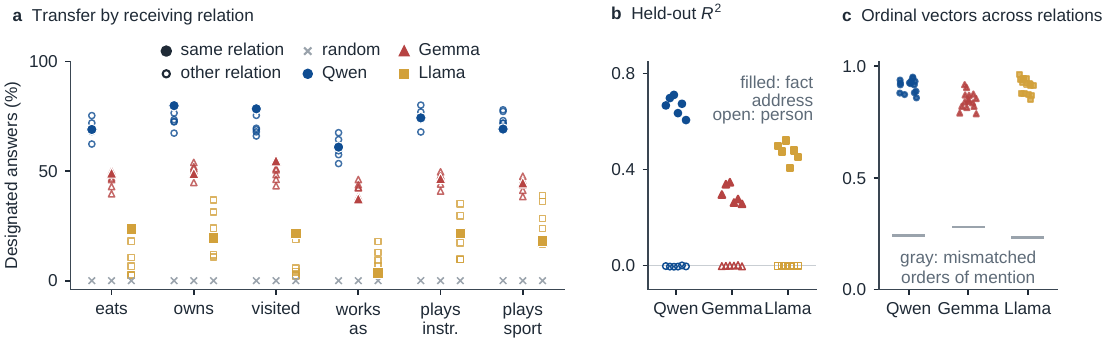}
\caption{\textbf{The same ordinal vector acts under every relation.} (a) Designated answers by receiving relation: an ordinal vector calibrated on the same relation (filled), on each of the five other relations at its own length (open), and random directions ($\times$); 64 test word sets, 3,072 cases per point. Not plotted: the \emph{eats} vector rescaled to the receiving relation's change length comes within five points of the filled marker for every relation (paired differences $-4.0$ to $+4.6$ points). (b) Held-out $R^2$ of query states under fact-address (filled) and person (open) templates, one point per relation. (c) Cosine between ordinal vectors fit separately on two relations, one point per relation pair, averaged over pairs of orders of mention; gray bars are the cosine between mismatched ordinal vectors.}
\label{fig:relation_generality}
\end{figure}

Unedited answers are correct in at least $97.5\%$ of cases for every relation and model, and the ceiling reaches $88$--$99\%$. Random directions select the designated fact in at most $0.03\%$ of cases. Averaged over relations, the receiving relation's own vector selects the designated fact in $72\%$, $47\%$, and $18\%$ of Qwen, Gemma, and Llama cases, and a vector from another relation in $70\%$, $46\%$, and $16\%$ (\cref{fig:relation_generality}a). Rescaled to the receiving relation's length, the \emph{eats} vector comes within five points of each relation's own vector in every model (paired differences from $-4.0$ to $+4.6$ points). Fact-address templates give held-out $R^2$ of $0.61$--$0.71$, $0.26$--$0.35$, and $0.41$--$0.52$, while person templates stay within $0.01$ of zero (\cref{fig:relation_generality}b). The ordinal vectors of two relations point the same way, with cosines of $0.79$--$0.96$ over relation pairs and models, against $0.18$--$0.38$ between mismatched ordinal vectors (\cref{fig:relation_generality}c); the address subspaces of the six relations nearly coincide as well (\cref{app:address_subspace}). The relation changes the contents at an order of mention, not the ordinal vector between them.

\subsection{Order of Mention versus Token Distance}
\label{app:address_controls}
The ordinal vector follows order of mention, not token distance (\cref{sec:selection}). When all facts have the same length, the $k$-th fact is also a fixed number of tokens before the question, so an address by order of mention cannot be told apart from one by distance. Each context of 64 new four-person word sets is rendered under eight length conditions with the same arrangement and content assignments: uniform facts; one fact lengthened by a neutral opening clause (four conditions); the second and fourth facts lengthened; filler lines after facts 1 and 3; and a filler paragraph before the question. Ordinal vectors are calibrated on uniform contexts from 16 other word sets, or on mixed-length contexts, and applied unchanged.

\begin{figure}[!ht]
\centering
\includegraphics[width=\linewidth]{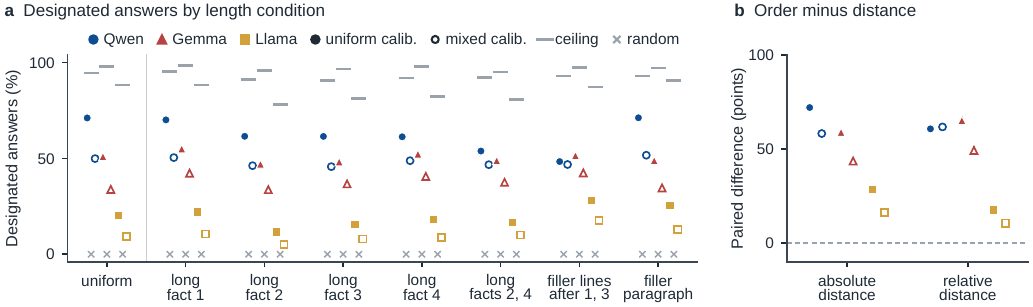}
\caption{\textbf{The ordinal vector follows order of mention, not token distance.} (a) Designated answers by length condition for ordinal vectors calibrated on uniform (filled) or mixed-length (open) contexts, with the ceiling (gray ticks) and random directions ($\times$). (b) Answers naming the fact at the calibrated order minus those naming the fact at the calibrated absolute or relative distance, on edits where the two facts differ from each other and from the unedited answer; markers as in (a), with 95\% intervals over 64 word sets.}
\label{fig:address_controls}
\end{figure}

For an edit from order of mention $p$ to $p'$ on a question about fact $s$, let $d(s)$ be the token distance of $s$ from the question and $d_{\mathrm{ref}}[p]$ the distance of order $p$ in uniform contexts. The \emph{absolute} distance address selects the fact whose distance is nearest $d_{\mathrm{ref}}[p']$; the \emph{relative} one selects the fact whose distance is nearest $d(s)+d_{\mathrm{ref}}[p']-d_{\mathrm{ref}}[p]$; ties go to the fact nearer the question. Both rules and the contrast were fixed before testing: answers naming the fact at the calibrated order minus answers naming the distance-predicted fact, on edits where the two facts differ from each other and from the fact the unedited question asks about.

Transfer survives every length condition, from $48\%$ to $71\%$ in Qwen, from $47\%$ to $55\%$ in Gemma, and from $12\%$ to $28\%$ in Llama, with random directions at zero (\cref{fig:address_controls}a). On edits where order and distance predict different facts, the ordinal vector selects the fact at the calibrated order far more often than the fact at the calibrated distance: by $72.1$ $[70.0, 74.0]$ points in Qwen, $58.9$ $[57.0, 60.9]$ in Gemma, and $28.4$ $[26.6, 30.1]$ in Llama against absolute distance, and by $60.7$, $65.2$, and $17.6$ points against relative distance. Mixed-length calibration gives the same verdict. Held-out query states are fit by fact-address templates ($R^2$ $0.52$, $0.26$, and $0.41$) and not by a token-distance template, the fact-address template of the order of mention whose mean token distance from the question in uniform-length contexts is nearest the queried fact's distance ($0.02$, $-0.11$, $-0.15$); both templates are fit on calibration word sets and scored on the test word sets. Transfer counts facts, not tokens.

%% file: appendix/address_subspace.tex
\subsection{The Address Subspace}
\label{app:address_subspace}
Fact addresses span a shared low-rank subspace (\cref{sec:address_geometry}): the three-dimensional span of the fact-address templates holds nearly as much of each held-out query change as the best three-dimensional subspace. The numbers of \cref{sec:address_geometry} use the main-block states of \cref{app:query_geometry}, with templates fit on all 16 calibration word sets; \cref{fig:address_linear}a, like \cref{fig:query_organization}, fits the templates on eight of them and shows the states of the other eight.

\paragraph{Share of held-out changes.}
With the fact-address templates $a_p$ of \cref{app:query_geometry} fit on the 16 calibration word sets, the address subspace is $A=\mathrm{span}\{a_p-a_1\}$, three-dimensional for four facts. For each of the 128 test contexts and each of its six pairs of orders of mention, the per-context change is $\Delta_C(p\to p')=h(C,p')-h(C,p)$, where $h(C,p)$ is the state for the question about the fact at order of mention $p$ (the reverse change has the same share). Its share inside $A$ is $\|P_A\Delta_C\|^2/\|\Delta_C\|^2$, averaged over the 768 changes: $69\%$, $33\%$, and $52\%$ in Qwen, Gemma, and Llama.

Two references bound this share. The best three-dimensional subspace for these changes, the top three principal directions of the held-out changes themselves, holds $70\%$, $41\%$, and $60\%$; this is an in-sample upper bound. One hundred random three-dimensional subspaces hold $0.1\%$ on average, as expected for dimensions 3,584, 2,560, and 4,096. Relative to the gap between random and best, $A$ reaches $99\%$, $81\%$, and $87\%$.

\paragraph{Across relations.}
For each relation $r$ of the six in \cref{app:relation_generality}, a separate subspace $A_r$ is fit on that relation's 16 calibration word sets at the main block. For two relations $s$ and $t$, the principal angles between $A_s$ and $A_t$ give three cosines. The smallest of them, the cosine of the largest principal angle, averaged over the 15 relation pairs, is $0.89$, $0.78$, and $0.88$ (ranges over pairs $0.83$--$0.94$, $0.66$--$0.90$, and $0.80$--$0.94$); for random three-dimensional subspaces, even the largest of the three cosines is only about $0.05$.

\paragraph{Composition on disjoint word sets.}
Write $v_S(p\to p')$ for the mean of $\Delta_C(p\to p')$ over the calibration contexts of a subset $S$ of word sets. The 16 calibration word sets are split into two halves, and one half into two disjoint quarters $S_1,S_2$. For each of the 24 ordered triples $p\to p'\to p''$ of distinct orders of mention, we compare the target $v_T(p\to p'')$ on the other half with the sum $v_{S_1}(p\to p')+v_{S_2}(p'\to p'')$, so the three vectors come from disjoint word sets and no identity links the sum to the target. Averaged over triples and over the assignment of halves and quarters, the cosine is $0.99$, $0.97$, and $0.98$, and the relative error $\|{\rm sum}-{\rm target}\|/\|{\rm target}\|$ is $0.15$, $0.23$, and $0.18$. The null relabels the summands over the 23 non-identity relabelings of the orders of mention $\{1,2,3,4\}$: cosine $-0.09$ and relative error $1.45$--$1.56$ in all three models. The target estimated directly on $S_1\cup S_2$ agrees with $v_T(p\to p'')$ at cosine $0.995$, $0.989$, and $0.995$, the noise ceiling. Ordinal vectors add along paths, as \cref{cor:linear} predicts.

\subsection{Longer Lists}
\label{app:longer_lists}
In longer lists, fact addresses form an ordered chain whose later links shorten, and transfer is highest to the first fact (\cref{sec:address_geometry}). The transfer test of \cref{sec:selection} is repeated with lists of four, six, and eight single-relation facts, one person per fact, in all three models at their main blocks. Each test word set has eight people and eight contents, and a list of length $N$ uses the first $N$ of each, so the three lengths share names and contents; within a word set every person and every content appears once at every order of mention. Calibration uses 16 word sets and every combination of content assignment and arrangement; all $N(N-1)$ directed pairs of orders of mention are calibrated and tested (12, 30, and 56) on 64 test word sets, each in its $N$ cyclic arrangements. This experiment draws its names and contents from a separate pool, on which four-fact transfer is lower than in \cref{tab:contextual_selection}A (Qwen $75.6\%$ versus $84.1\%$).

\begin{figure}[!ht]
\centering
\includegraphics[width=\linewidth]{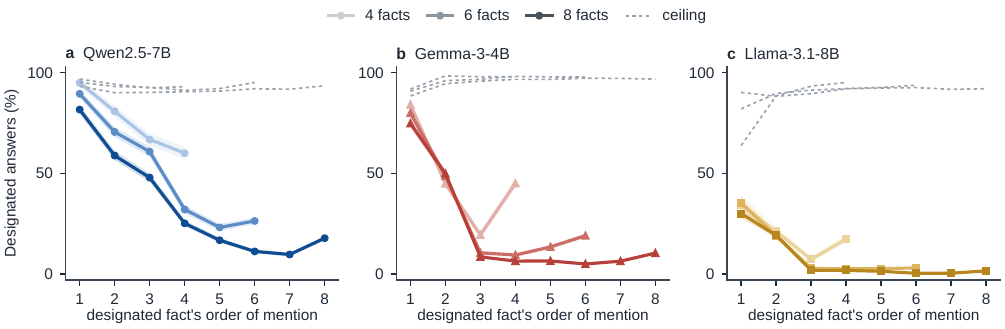}
\caption{\textbf{Transfer falls with the designated fact's order of mention and is highest for the first fact.} Designated answers for the ordinal vector by the designated fact's order of mention, in lists of four, six, and eight facts, with word-set bootstrap 95\% intervals (bands, mostly narrower than the markers); dashed lines are the ceiling. Random directions give at most $0.1\%$ at every length.}
\label{fig:longer_lists}
\end{figure}

\Cref{fig:longer_lists} shows two regularities. First, the ordinal vector transfers at every length, far above random directions: overall $75.6$, $50.4$, and $33.7\%$ in Qwen for four, six, and eight facts, $48.4$, $30.2$, and $21.1\%$ in Gemma, and $20.1$, $11.0$, and $7.2\%$ in Llama, while the ceiling stays high at every length (\cref{fig:longer_lists}). Second, transfer depends on the designated fact's order of mention: it is highest for the first fact (Qwen $82$--$95\%$), falls toward the middle, and rises again at the last fact in eight of the nine lists, a serial-position curve with a strong primacy effect \citep{henson1998startend}.

\paragraph{The chain of addresses.}
For lists of $N=4,6,8$ facts, the templates $a_1,\dots,a_N$ average the query-dependent states of the 16 calibration word sets, over all $N$ arrangements and $N$ content assignments, by the order of mention of the queried fact; the templates from two halves of these word sets have mean cosine at least $0.98$. Lengths are in units of $\|a_2-a_1\|$ at each $N$. \Cref{fig:address_linear}b projects the eight templates for $N=8$ on their top two principal directions, which hold $58\%$, $84\%$, and $77\%$ of their variance. \Cref{fig:address_linear}c plots $\|a_{p+1}-a_p\|$; for $N=8$ the last link is $0.44$, $0.13$, and $0.25$ of the first. \Cref{fig:address_linear}d plots, for each model, length $N$, and order of mention $p$ (54 points), the distance $\min_{p'\ne p}\|a_p-a_{p'}\|$ from $a_p$ to its nearest other address against the transfer rate to the fact at $p$ on the test word sets (\cref{fig:longer_lists}). Subtracting each model's mean distance and mean rate before the least-squares fit gives $R^2=0.82$ (Spearman $0.89$); without centering, $R^2=0.78$.

\subsection{Prefixes of a List}
\label{app:prefixes}
Each fact address is fixed once its fact is read: removing later facts leaves earlier addresses in place (\cref{sec:address_geometry}). Every eight-fact context of \cref{app:longer_lists} (the 16 calibration and 64 test word sets, eight contexts each: the eight cyclic arrangements, each with one content assignment that varies across word sets) is cut after its first $k$ facts, $k=2,\dots,8$, and asked about each of its $k$ facts, with states at the main blocks. Templates $a^{(k)}_p$ are computed within each prefix length from all 80 word sets as in \cref{app:query_geometry} and compared only through their differences, which do not depend on how the query-dependent states are centered.

For facts before the prefix's last, the vector from fact 1 to fact $p$ has median cosine $0.99$ with its eight-fact counterpart in every model (minimum $0.95$), and its length changes by at most $8\%$. The gap between neighboring addresses, relative to the first gap, depends on the fact's order of mention and not on the prefix length: in Gemma, gaps 2 to 4 are $0.62$--$0.66$, $0.41$--$0.43$, and $0.18$--$0.20$ at every $k$, and the first gap is constant in absolute size. Gap sizes indexed by order of mention, fit on the other prefix lengths, predict the gaps of a held-out prefix length with $R^2=0.97$, $0.99$, and $0.91$ in Qwen, Gemma, and Llama; indexed by distance from the list's end, $R^2<0$. The last fact of a prefix deviates slightly (median cosine $0.91$, $0.97$, and $0.94$), and in Llama its gap is larger by $0.075$ of the first gap, a recency effect.

%% file: appendix/composition.tex
\subsection{Person and Relation Do Not Add}
\label{app:composition}
Fact addresses predict that a person change and a relation change need not add up to the joint change, and they predict the direction of the gap (\cref{sec:address_geometry}). Contexts state four facts about two people and two relations, \emph{Alice eats the apple. Alice carries the pear. Bob eats the plum. Bob carries the peach.} Index the question cells (\cref{app:rearranged}) by person and relation, $(e,r)\in\{0,1\}^2$, and let $p_o(e,r)$ be their fact's order of mention under arrangement $o$.
\begin{lemma}[Fact addresses induce a person--relation interaction]
\label{lem:position_interaction}
Under \cref{hyp:conditional_reuse} with $\varepsilon=0$, the query state of the question about cell $(e,r)$ is $h_{er}=h_\ell(C,q_{er})=c_C+a_{p_o(e,r)}$, and the four-cell contrast of these states is
\begin{equation}
I=h_{11}-h_{10}-h_{01}+h_{00}=a_{p_o(11)}-a_{p_o(10)}-a_{p_o(01)}+a_{p_o(00)}.
\label{eq:position_interaction}
\end{equation}
The right-hand side need not vanish and can change with the arrangement.
\end{lemma}
\begin{proof}
Substitute $h_{er}=c_C+a_{p_o(e,r)}$ into the four-cell contrast. The context term has coefficients $+1-1-1+1=0$, leaving \cref{eq:position_interaction}. For a scalar example, take $a_1=0$, $a_2=1$, $a_3=2$, $a_4=4$, and place the cells $00,01,10,11$ at orders of mention $1,2,3,4$. The contrast is $4-2-1+0=1$. Exchanging the orders of mention of the last two cells changes it to $2-4-1+0=-3$, although neither case has a separate person-by-relation term.
\end{proof}

\paragraph{Joint versus summed changes.}
The contexts above appear in four arrangements: grouped by person, grouped by relation, and their reversals. Calibration uses 16 word sets, and each change is the calibration mean for its pair of question cells within the same arrangement. From a starting cell, $v_E$ changes the person, $v_R$ the relation, and $v_J$ both; the \emph{same-start sum} is $v_E+v_R$, with both terms measured from the starting cell. Each of 64 test word sets contributes four arrangements, content assignments, and starting cells each: 4,096 cases per condition, all scored against the joint target.

\begin{table}[!ht]
\centering\small
\caption{\textbf{The joint change selects the joint target more often than the sum of its parts.} Designated answers (\%), 64 test word sets, 4,096 cases per edit; $\pm$: half the 95\% paired word-set bootstrap interval, in points; bold: the joint change.}
\label{tab:composition_selection}
\vspace{4pt}
\begin{tabular}{lrrr}
\toprule
Edit & Qwen & Gemma & Llama\\
\midrule
Joint change $v_J$ & \textbf{68.5} & \textbf{57.1} & \textbf{26.0}\\
Same-start sum $v_E+v_R$ & 43.8 & 44.0 & 18.8\\
Sum rescaled to $\|v_J\|$ & 35.3 & 31.7 & 14.1\\
Ceiling (own state for the joint question) & 87.9 & 90.7 & 56.7\\
\midrule
Joint minus sum & 24.7{\,\tiny$\pm$1.3} & 13.1{\,\tiny$\pm$1.4} & 7.2{\,\tiny$\pm$0.9}\\
Joint minus rescaled sum & 33.2{\,\tiny$\pm$1.5} & 25.4{\,\tiny$\pm$1.4} & 11.9{\,\tiny$\pm$1.0}\\
\bottomrule
\end{tabular}
\end{table}

The sum is longer than the joint change, so its shortfall is not a lack of length: rescaling it to $\|v_J\|$ lowers its rate further (\cref{tab:composition_selection}). The person and relation changes are nearly orthogonal (cosine $0.11$, $0.13$, and $0.12$), and their sum has cosine $0.88$, $0.86$, and $0.89$ with $v_J$; the interaction $I=v_J-v_E-v_R$, the four-cell contrast of \cref{lem:position_interaction}, has $54\%$, $59\%$, and $51\%$ of the norm of $v_J$.

\begin{figure}[!ht]
\centering
\includegraphics[width=0.42\linewidth]{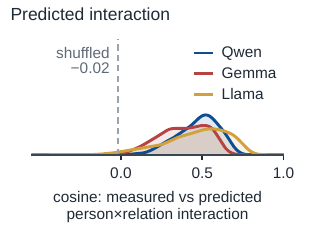}
\caption{\textbf{Fact addresses predict the person--relation interaction.} Distribution, over 3,328 held-out contexts per model, of the cosine between the measured interaction $I$ of the query-dependent states (\cref{eq:position_interaction}) and its prediction from fact-address templates and the context's orders of mention; means $0.47$, $0.38$, and $0.46$ in Qwen, Gemma, and Llama. Dashed line: the mean cosine when the order labels are shuffled ($-0.02$).}
\label{fig:query_controls}
\end{figure}

\paragraph{The interaction predicted by fact addresses.}
For the prediction of \cref{lem:position_interaction}, fact-address templates are fit on person-by-relation contexts from 16 calibration word sets, rendered in the twelve even permutations of the four facts with four content assignments. Held-out contexts are 64 test word sets rendered in the twelve odd permutations, which calibration never used, and in the original arrangement (an even permutation, also used in calibration; its word sets are held out), with four content assignments: 3,328 contexts. In each, the measured interaction $I=y_{11}-y_{10}-y_{01}+y_{00}$ of the query-dependent states is compared with $a_{p_o(11)}-a_{p_o(10)}-a_{p_o(01)}+a_{p_o(00)}$ computed from the templates and the orders of mention of the four cells in that context's arrangement. The mean cosine is $0.47$, $0.38$, and $0.46$ at the main blocks (\cref{fig:query_controls}). Relabeling the templates by each of the 23 non-identity relabelings of the orders of mention gives $-0.02$ in every model, and at the final block the cosine falls to $0.07$, $0.04$, and $0.07$. The measured interaction points toward the one fact addresses predict, far above relabeled orders, and the agreement vanishes at the output.

%% file: appendix/depth.tex
\subsection{The Depth Window}
\label{app:all_block_donors}
The fact address is shared as a difference only within a depth window, after which whole states still carry it and then the donor's content takes over (\cref{sec:transfer_depth}). The depth curves in \cref{fig:selection_stages}b--d repeat the donor design of \cref{app:transfer_facts_donors} at every block of all three models, on the prompts of \cref{tab:contextual_selection}C: the 16 calibration word sets as donors, crossed with the 64 test word sets as receivers in one fixed arrangement, two content assignments, and four directed person changes. At every block the conditions are the unedited question, the ordinal vector re-estimated at that block from the calibration word sets, the ceiling, and the donor's state for the other question; the main and final blocks add the donor's state for the same question (\cref{tab:contextual_selection}C). Intervals for donor conditions resample donors and receivers independently; intervals for the other conditions resample receivers.

\paragraph{The example.}
\Cref{fig:selection_stages}a shows the first case of Gemma's block-selection word sets: a receiver asked about its first fact (\emph{apple crumble}) takes the state of a donor asked about its second (\emph{baked trout}), and answers \emph{pear tart}, its own second fact, at block 22, but \emph{baked apple crumble}, the donor's first word followed by its own content, at block 33.

\paragraph{Scoring donor outputs.}
Each donor output is scored three ways: it names the receiver's designated content, it is the donor's own complete answer, or it begins with the first token of the donor's answer. Past the window most outputs begin with the donor's first word and continue with receiver content, as in \emph{baked apple crumble}; such outputs name no content of either context and count as invalid. First-token agreement therefore measures when the donor's content takes over, and \cref{fig:selection_stages} reports it alongside complete answers. Complete donor answers remain rare at every block, at most $2.1\%$ in Qwen, $9.8\%$ in Gemma, and $4.1\%$ in Llama.

\begin{figure}[!ht]
\centering
\includegraphics[width=\linewidth]{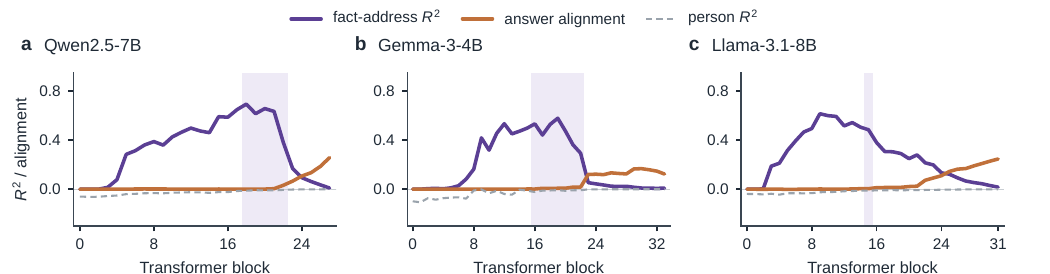}
\caption{\textbf{The fact-address geometry follows the stages of \cref{fig:selection_stages}.} Query-state geometry at every block: $R^2$ of fact-address and person templates (leave-one-word-set-out over the 16 calibration word sets, 256 contexts), and each query state's alignment with its answer's first-token output direction (128 test contexts). Shading: the depth window of \cref{fig:selection_stages}.}
\label{fig:depth_geometry}
\end{figure}

\paragraph{Answer alignment.}
For the alignment curves of \cref{fig:depth_geometry}, the output direction of an answer in a context is the final-norm scale times the unembedding row of the answer's first token, and the four directions of a context are centered. Alignment is the cosine between each query-dependent state and its own answer's direction, averaged over test contexts; it uses only the unembedding and final-norm weights. It is near zero at the main blocks ($0.001$, $0.017$, and $0.004$) and reaches $0.26$, $0.12$, and $0.25$ at the final block. Past the window, states align with their own answer's output direction.

%% file: appendix/training.tex
\subsection{Pretraining Checkpoints}
\label{app:development}
In pretraining, questions switch from the named person to order of mention before the model can use the address, and the depth profile settles last (\cref{sec:transfer_depth}). \Cref{fig:development} repeats the tests on ten checkpoints of OLMo-2-7B \citep{olmo2024olmo2}: stage-1 steps 1k, 4k, 10k, 25k, 50k, 99k, 200k, 400k, and 929k (5B to 3.9T training tokens) and the released model after stage-2 mid-training. The prompt is the plain format of \cref{app:base_instruct} with the tokenizer's \texttt{<|endoftext|>} as the first token, which marks a document start in OLMo-2's packed pretraining data. Each checkpoint runs the protocol of \cref{app:scale} with the depth curves: its own rule-chosen block, the transfer test, the cosines, the query-state geometry at every block, and the depth curves at every second block (8 donors $\times$ 32 receivers). Transfer and cosines are read at block 16, the released model's rule-chosen block; checkpoints whose rule-chosen block differs are also evaluated at block 16. Early checkpoints answer in sentences, so answers are scored by the single stated content they name. A base checkpoint can name the designated answer without any edit (up to $14\%$ of cases), so the stages of \cref{fig:development}a apply the thresholds of \cref{fig:selection_stages}a to the gain over the unedited answer.

\input{figures/development-table}

\Cref{tab:development} gives the numbers behind \cref{fig:development}b,c. At 5B tokens the reversed-order cosine is $+0.98$: the change for a pair of people does not depend on their order, as expected if questions were represented by the named people. It falls to $0.25$ at 17B and $-0.61$ at 42B. Fact-address geometry precedes the ability to answer: at 17B tokens the model names the right content in $24.9\%$ of cases, the rate of guessing among four, yet fact-address templates already explain $0.29$ of the variance.

The depth profile moves throughout training (\cref{fig:development}a). Shared-address blocks appear at 42B, vanish at 105B, and are present from 210B through 1.7T; content blocks appear from 416B. All three stages first co-occur at 839B and again at 1.7T, each time separated by blocks that reach no threshold. At 3.9T whole-state and content blocks remain, but no block reaches the shared-address threshold, and only the released model shows all three stages as one contiguous profile. Read at the fixed block 16, transfer therefore peaks at 210B ($50.1\%$ against a ceiling of $56.8\%$) and falls later as the profile moves: the ceiling at block 16 drops from $55.9\%$ at 416B to $24.3$--$35.9\%$ from 839B to 3.9T and returns to $61.8\%$ after mid-training. Questions are organized by order of mention early in pretraining; the depth profile settles only late.

%% file: figures/development-table.tex
\begin{table}[!ht]
\centering\small
\caption{\textbf{OLMo-2-7B checkpoints.} Rule-chosen block; at block 16: unedited accuracy and designated answers (\%) for the ordinal vector (bold), a random direction and the ceiling, scored by the single stated content an answer names, 64 test word sets and 12,288 cases per edit; new-name and reversed-order cosines of \cref{prop:invariance}; and the peak leave-one-word-set-out fact-address $R^2$ over blocks (its block). \emph{final}: the released model after stage-2 mid-training.}
\label{tab:development}
\vspace{4pt}
\setlength{\tabcolsep}{4pt}
\begin{tabular}{@{}lrrrrrrrr@{}}
\toprule
& & \multicolumn{6}{c}{At block 16} & \\
\cmidrule(lr){3-8}
& & \multicolumn{4}{c}{Scored by stated content (\%)} & \multicolumn{2}{c}{Cosine} & \\
\cmidrule(lr){3-6}\cmidrule(lr){7-8}
Tokens & \shortstack{Rule\\block} & Unedited & $v$ & Random & Ceiling & New names & Reversed & Peak $R^2$ (block)\\
\midrule
5B & 22 & 1.3 & \textbf{1.2} & 1.2 & 1.2 & 0.09 & 0.98 & 0.00 (31)\\
17B & 22 & 24.9 & \textbf{16.3} & 16.0 & 16.7 & 0.37 & 0.25 & 0.29 (11)\\
42B & 14 & 50.0 & \textbf{25.1} & 14.6 & 31.5 & 0.70 & $-$0.61 & 0.52 (13)\\
105B & 22 & 63.9 & \textbf{23.7} & 11.3 & 27.6 & 0.76 & $-$0.74 & 0.61 (13)\\
210B & 16 & 67.4 & \textbf{50.1} & 9.3 & 56.8 & 0.79 & $-$0.74 & 0.74 (13)\\
416B & 16 & 79.9 & \textbf{46.0} & 6.3 & 55.9 & 0.80 & $-$0.71 & 0.71 (13)\\
839B & 16 & 86.1 & \textbf{21.7} & 4.6 & 29.7 & 0.75 & $-$0.56 & 0.62 (13)\\
1.7T & 16 & 91.5 & \textbf{23.5} & 2.0 & 35.9 & 0.80 & $-$0.65 & 0.69 (14)\\
3.9T & 22 & 89.2 & \textbf{14.6} & 2.5 & 24.3 & 0.77 & $-$0.59 & 0.66 (13)\\
final & 16 & 92.4 & \textbf{33.1} & 1.8 & 61.8 & 0.85 & $-$0.78 & 0.71 (14)\\
\bottomrule
\end{tabular}
\end{table}

%% file: appendix/generality.tex
The main text reports its tests on three instruction-tuned models and templated lists, and states that they hold from 1.5B to 32B parameters (\cref{tab:scale}), in base weights (\cref{fig:relational_hypothesis}d), and in short paragraphs (\cref{sec:selection}); this section gives those tests on eleven further models and on prose: the instruction-tuned Qwen2.5-1.5B, Qwen2.5-3B, Qwen2.5-14B, Qwen2.5-32B, Qwen3-8B, and Llama-3.2-3B, and the base weights of Qwen2.5-1.5B, Qwen2.5-3B, Qwen3-8B, Gemma-3-4B, and Llama-3.2-3B. Materials, calibration and test word sets, and scoring follow \cref{app:selection_protocol} except where noted, and every rate is a share of complete greedy answers.

\subsection{Model Scale}
\label{app:scale}
Ordinal addressing holds at every model size tested (\cref{sec:selection,sec:transfer_depth}). We run Qwen2.5-Instruct at 1.5B, 3B, 14B, and 32B parameters and Qwen3-8B with its thinking mode switched off. Each model's main block follows the rule of \cref{app:selection_protocol}, and the 7B model keeps its block 20. Each model then runs the transfer test of \cref{tab:contextual_selection}A, the cosines of \cref{fig:relational_hypothesis}d, the templates of \cref{app:query_geometry}, and the depth curves of \cref{fig:selection_stages} with 8 donors and 32 receivers.

Every model addresses facts by order of mention (\cref{tab:scale}). New names keep the direction of the ordinal vector and reversed orders reverse it, with cosines of $0.92$ to $0.97$ and $-0.85$ to $-0.96$; at the main block, the fact-address template explains $0.46$--$0.76$ of the query-dependent variance (leave-one-word-set-out) and person none. The ordinal vector selects the designated fact in $34.6$--$84.1\%$ of cases and random directions almost never. Its rate does not rise with size, and neither does the ceiling: the ceiling selects $65.2$--$96.9\%$ of designated facts, lowest in the 3B and 14B models, and in the 3B model the chat template lowers it (\cref{app:base_instruct}). Relative to the ceiling, the ordinal vector reaches $68$--$87\%$ in Qwen2.5 and $46\%$ in Qwen3-8B. In every model the ordinal vector acts within a window of a few blocks at $63$--$86\%$ of relative depth (block divided by the last block; the window is where it selects at least $10\%$, as in \cref{fig:selection_stages}), and whole donor states outlast it: two blocks past the window they still select $7$--$30\%$ of receiver facts while the ordinal vector selects under $5\%$, as in \cref{sec:transfer_depth}.

\input{figures/generality-scale-table}

\subsection{Base and Instruction-Tuned Weights}
\label{app:base_instruct}
Instruction tuning is not needed for ordinal addressing (\cref{sec:geometric_hypothesis}). We compare five base models with their instruction-tuned partners on one plain prompt, the instruction, facts, and question followed by \texttt{Answer:}, and add each instruct model in its native chat template. Each row chooses its own main block by the rule above.

In Qwen the base weights carry the address about as strongly as the tuned ones; in Llama the base model is weaker (\cref{tab:base_instruct}). On the identical prompt, the ordinal vector selects $67.3\%$ against $69.6\%$ in Qwen2.5-1.5B, $80.0\%$ against $72.2\%$ in Qwen2.5-3B, $31.6\%$ against $33.3\%$ in Qwen3-8B, and $41.9\%$ against $54.8\%$ in Llama-3.2-3B. The four Qwen and Llama base models keep a new-name cosine of at least $0.86$ and a reversed-order cosine of at most $-0.74$, with fact-address $R^2$ between $0.48$ and $0.78$. Gemma-3-4B's base model answers in full sentences, so we score the single stated content each answer names: it answers $92.9\%$ of unedited questions correctly, and the ordinal vector selects the designated fact in $21.9\%$ of cases against $0.6\%$ for random directions, below its instruct partner; its new-name and reversed-order cosines, $0.76$ and $-0.44$, are also the weakest, while its fact-address $R^2$, $0.49$, equals that partner's on the same prompt.

Within Qwen, the one large gap comes from the chat template, not the tuning: Qwen2.5-3B-Instruct selects $72.2\%$ with the plain prompt and $48.0\%$ in its chat template, where the ceiling also falls from $90.3\%$ to $66.1\%$. The prompt format, not the tuning, changes how strongly the ordinal vector acts.

\input{figures/generality-base-table}

\subsection{Facts Stated in Prose}
\label{app:prose}
Fact addresses carry over from lists to prose (\cref{sec:selection}). Each prose context is one paragraph stating the four facts with four different verb frames drawn from sixteen, such as \emph{For lunch, Petra had the buckwheat} or \emph{The teff ended up in front of Sabrina}; none uses \emph{eats}, and four frames put the content first. Two or three unrelated sentences, such as \emph{It had rained earlier that morning, and the streets were still wet}, sit between facts, and an opening or closing narrative sentence is added with probability one half. A fact's order of mention is its place in the paragraph. Paragraphs run 37--87 words, against 20--30 for the list, and the instruction and question are those of the main text. Test word sets are 64 new word sets whose names and contents appear in no main-text word set; each is also rendered as a list with the same people, contents, and questions.

\input{figures/generality-prose-table}

The ordinal vector calibrated on the main-text lists, unchanged, selects the designated fact in paragraphs (\cref{tab:prose}): $44.5\%$, $42.4\%$, and $14.0\%$ of cases in Qwen, Gemma, and Llama, against $80.5\%$, $52.2\%$, and $29.3\%$ on the matched lists, while random directions select none. It stays above random at every designated order of mention (lowest $22.9\%$, $15.0\%$, and $3.6\%$), and the ceiling still reaches $92.8\%$, $96.3\%$, and $83.1\%$. To separate format from word sets, 16 new word sets are rendered both as paragraphs and as lists for calibration. On prose receivers, the ordinal vector calibrated on their paragraphs selects fewer designated facts than the one calibrated on their lists ($40.2\%$ against $44.0\%$ in Qwen, $32.3\%$ against $43.4\%$ in Gemma, $8.5\%$ against $10.2\%$ in Llama), and on list receivers the paragraph-calibrated vector still selects $41.1\%$, $24.1\%$, and $12.2\%$: the drop on prose comes from the receivers, not from a vector specific to lists.

The geometry agrees. In paragraphs, the fact-address template explains $0.45$, $0.23$, and $0.33$ of the query-dependent variance at the main block (leave-one-word-set-out) and person none. Fact-address templates from prose and from lists point the same way (cosines $0.76$, $0.81$, and $0.84$), as do their ordinal vectors ($0.75$, $0.80$, and $0.81$), and $35\%$, $17\%$, and $34\%$ of the squared norm of query-dependent prose states lies in the three-dimensional span of the main-text ordinal vectors, against $0.1\%$ for random subspaces. Paragraphs largely reuse the fact addresses of lists, though less of each prose state lies in them.

%% file: figures/generality-scale-table.tex
\begin{table}[h]
\centering\footnotesize
\setlength{\tabcolsep}{2.6pt}
\caption{\textbf{Ordinal addressing holds from 1.5B to 32B parameters.} Every model runs the transfer test of \cref{tab:contextual_selection}A (64 new word sets, 12,288 cases), the cosines of \cref{fig:relational_hypothesis}d, the fact-address and person templates of \cref{app:query_geometry}, and the depth curves of \cref{fig:selection_stages} at every second block. Main block: block index / number of blocks, chosen on block-selection word sets by the rule of \cref{app:selection_protocol}. Unedited: correct answers without an edit. Window: first to last block at which the ordinal vector selects at least $10\%$, as in \cref{fig:selection_stages} (8 donors $\times$ 32 receivers). $^\dagger$The main-text model; its block was chosen before this sweep (\cref{app:selection_protocol}), and its window comes from the every-block curves of \cref{fig:selection_stages}.}
\label{tab:scale}
\vspace{6pt}
\begin{tabular}{@{}lrrrrrrrrrr@{}}
\toprule
& & \multicolumn{4}{c}{Designated answers (\%)} & \multicolumn{2}{c}{Change cosine} & \multicolumn{2}{c}{$R^2$} & \\
\cmidrule(lr){3-6}\cmidrule(lr){7-8}\cmidrule(lr){9-10}
Model & \shortstack{Main\\block} & Unedited & $v$ & Random & Ceiling & New names & Reversed & \shortstack{Fact\\address} & Person & Window\\
\midrule
Qwen2.5-1.5B & 18/28 & 98.6 & \textbf{72.0} & 0.1 & 95.7 & 0.97 & $-$0.96 & 0.76 & $-$0.02 & 18--22\\
Qwen2.5-3B & 26/36 & 96.6 & \textbf{48.0} & 0.0 & 66.1 & 0.92 & $-$0.85 & 0.65 & $-$0.02 & 26--30\\
Qwen2.5-7B$^\dagger$ & 20/28 & 99.9 & \textbf{84.1} & 0.0 & 96.9 & 0.93 & $-$0.93 & 0.66 & $-$0.01 & 18--22\\
Qwen2.5-14B & 34/48 & 99.9 & \textbf{44.4} & 0.0 & 65.2 & 0.92 & $-$0.92 & 0.46 & $-$0.01 & 32--34\\
Qwen2.5-32B & 48/64 & 100.0 & \textbf{55.1} & 0.0 & 74.7 & 0.94 & $-$0.94 & 0.51 & $-$0.01 & 46--50\\
\addlinespace[2pt]
Qwen3-8B & 22/36 & 99.6 & \textbf{34.6} & 0.0 & 74.8 & 0.95 & $-$0.93 & 0.71 & $-$0.01 & 22--28\\
\bottomrule
\end{tabular}

\end{table}

%% file: figures/generality-base-table.tex
\begin{table}[h]
\centering\footnotesize
\setlength{\tabcolsep}{2.3pt}
\caption{\textbf{Base models address facts the same way.} Plain: the instruction, facts, and question followed by \texttt{Answer:}, with no chat template, identical for base and instruct weights; chat: the instruct model's native template. Each row chooses its own main block by the rule of \cref{app:selection_protocol} and runs the tests of \cref{tab:scale}. $^\ast$Gemma-3-4B base answers in sentences, so its answers are rescored by the single stated content they name. $^\ddagger$Main-text Gemma run (\cref{tab:contextual_selection}A, \cref{fig:relational_hypothesis}d, \cref{app:query_geometry}).}
\label{tab:base_instruct}
\vspace{6pt}
\begin{tabular}{@{}llrrrrrrrr@{}}
\toprule
& & & \multicolumn{4}{c}{Designated answers (\%)} & \multicolumn{2}{c}{Change cosine} & \\
\cmidrule(lr){4-7}\cmidrule(lr){8-9}
Model & Weights, prompt & Block & Unedited & $v$ & Random & Ceiling & New names & Reversed & \shortstack{Fact-\\address $R^2$}\\
\midrule
Qwen2.5-1.5B & base, plain & 18 & 87.7 & \textbf{67.3} & 1.7 & 82.0 & 0.97 & $-$0.95 & 0.78\\
 & instruct, plain & 20 & 98.7 & \textbf{69.6} & 0.1 & 95.3 & 0.95 & $-$0.94 & 0.63\\
 & instruct, chat & 18 & 98.6 & \textbf{72.0} & 0.1 & 95.7 & 0.97 & $-$0.96 & 0.76\\
\addlinespace[2pt]
Qwen2.5-3B & base, plain & 26 & 95.5 & \textbf{80.0} & 0.1 & 94.1 & 0.94 & $-$0.87 & 0.69\\
 & instruct, plain & 26 & 95.7 & \textbf{72.2} & 0.0 & 90.3 & 0.93 & $-$0.88 & 0.65\\
 & instruct, chat & 26 & 96.6 & \textbf{48.0} & 0.0 & 66.1 & 0.92 & $-$0.85 & 0.65\\
\addlinespace[2pt]
Qwen3-8B & base, plain & 22 & 96.9 & \textbf{31.6} & 0.0 & 71.7 & 0.95 & $-$0.93 & 0.64\\
 & instruct, plain & 22 & 98.5 & \textbf{33.3} & 0.0 & 74.6 & 0.92 & $-$0.89 & 0.64\\
 & instruct, chat & 22 & 99.6 & \textbf{34.6} & 0.0 & 74.8 & 0.95 & $-$0.93 & 0.71\\
\addlinespace[2pt]
Gemma-3-4B & base, plain & 22 & 6.7 & \textbf{14.0} & 0.1 & 27.6 & 0.76 & $-$0.44 & 0.49\\
 & \quad by content\rlap{$^\ast$} & & 92.9 & \textbf{21.9} & 0.6 & 52.6 & & & \\
 & instruct, plain & 20 & 94.4 & \textbf{47.7} & 0.0 & 85.7 & 0.93 & $-$0.88 & 0.49\\
 & instruct, chat\rlap{$^\ddagger$} & 22 & 99.9 & \textbf{53.9} & 0.0 & 96.4 & 0.88 & $-$0.82 & 0.29\\
\addlinespace[2pt]
Llama-3.2-3B & base, plain & 14 & 88.1 & \textbf{41.9} & 0.4 & 85.4 & 0.86 & $-$0.74 & 0.48\\
 & instruct, plain & 14 & 95.5 & \textbf{54.8} & 0.0 & 94.0 & 0.89 & $-$0.82 & 0.48\\
 & instruct, chat & 14 & 97.4 & \textbf{51.5} & 0.0 & 93.9 & 0.91 & $-$0.85 & 0.51\\
\bottomrule
\end{tabular}

\end{table}

%% file: figures/generality-prose-table.tex
\begin{table}[!ht]
\centering\footnotesize
\setlength{\tabcolsep}{4pt}
\caption{\textbf{Ordinal addressing carries over to prose.} Top: designated answers (\%) on 64 new word sets, 12,288 cases per entry, each edit added once at the main block. Receivers state the four facts either as a paragraph (\emph{prose}) or as the list of the main text, with the same people, contents, arrangements, and questions. The ordinal vectors are calibrated on the main-text lists (the $v$ of \cref{tab:contextual_selection}A), on lists of 16 new word sets, or on paragraphs of the same 16 word sets. Bottom: at the main block, leave-one-word-set-out $R^2$ of fact-address and person templates on paragraphs and on the matched lists (the list $R^2$ uses this experiment's word sets, not those of \cref{app:query_geometry}); mean cosine between matching prose and list fact-address templates and between matching prose and list ordinal vectors; fraction of the squared norm of query-dependent prose states inside the span of the main-text ordinal vectors, against random subspaces of the same dimension.}
\label{tab:prose}
\vspace{6pt}
\begin{tabular}{@{}llrrrrrr@{}}
\toprule
& & & \multicolumn{3}{c}{Ordinal vector calibrated on} & & \\
\cmidrule(lr){4-6}
Model & \shortstack{Receiver\\format} & Unedited & Lists (main) & Lists (new word sets) & Prose & Random & Ceiling\\
\midrule
Qwen & prose & 99.7 & \textbf{44.5}{\,\tiny$\pm$1.8} & 44.0{\,\tiny$\pm$1.8} & 40.2{\,\tiny$\pm$1.5} & 0.0 & 92.8{\,\tiny$\pm$1.6}\\
 & list & 99.9 & \textbf{80.5}{\,\tiny$\pm$2.1} & 78.1{\,\tiny$\pm$2.2} & 41.1{\,\tiny$\pm$1.7} & 0.0 & 95.4{\,\tiny$\pm$1.5}\\
\addlinespace[2pt]
Gemma & prose & 99.2 & \textbf{42.4}{\,\tiny$\pm$1.2} & 43.4{\,\tiny$\pm$1.2} & 32.3{\,\tiny$\pm$1.2} & 0.0 & 96.3{\,\tiny$\pm$0.9}\\
 & list & 99.7 & \textbf{52.2}{\,\tiny$\pm$1.4} & 52.8{\,\tiny$\pm$1.4} & 24.1{\,\tiny$\pm$1.2} & 0.0 & 96.5{\,\tiny$\pm$0.9}\\
\addlinespace[2pt]
Llama & prose & 99.6 & \textbf{14.0}{\,\tiny$\pm$1.1} & 10.2{\,\tiny$\pm$0.9} & 8.5{\,\tiny$\pm$0.9} & 0.0 & 83.1{\,\tiny$\pm$1.5}\\
 & list & 99.8 & \textbf{29.3}{\,\tiny$\pm$1.9} & 22.4{\,\tiny$\pm$1.7} & 12.2{\,\tiny$\pm$1.1} & 0.0 & 85.6{\,\tiny$\pm$1.7}\\
\bottomrule
\end{tabular}

\vspace{6pt}
\begin{tabular}{@{}lrrrrrrr@{}}
\toprule
& \multicolumn{2}{c}{Prose $R^2$} & List $R^2$ & \multicolumn{2}{c}{Prose vs.\ list} & \multicolumn{2}{c}{Squared norm in list span (\%)} \\
\cmidrule(lr){2-3}\cmidrule(lr){4-4}\cmidrule(lr){5-6}\cmidrule(lr){7-8}
Model & \shortstack{Fact\\address} & Person & \shortstack{Fact\\address} & Template cos. & Vector cos. & Prose states & Random\\
\midrule
Qwen & 0.45 & $-$0.01 & 0.71 & 0.76 & 0.75 & 35 & 0.1\\
Gemma & 0.23 & $-$0.01 & 0.39 & 0.81 & 0.80 & 17 & 0.1\\
Llama & 0.33 & $-$0.01 & 0.53 & 0.84 & 0.81 & 34 & 0.1\\
\bottomrule
\end{tabular}

\end{table}